\documentclass[review]{elsarticle}

\usepackage{lineno}
\modulolinenumbers[5]
\usepackage{footnote}
\usepackage{latexsym,bm,amsmath,amssymb} %Latex中数学包
\usepackage{graphicx}
\usepackage{algorithm}  
\usepackage{algorithmicx}  
\usepackage{algpseudocode}
\usepackage{subfigure}
\usepackage{multirow}
\usepackage{longtable}
\usepackage{relsize}
\usepackage{appendix}
\usepackage{diagbox,graphicx}
\usepackage{slashbox}
\usepackage{adjustbox}
\usepackage{supertabular,booktabs}
\usepackage{makecell}
\usepackage{array}
\usepackage[colorlinks,
linkcolor=black,
anchorcolor=black,
citecolor=black]{hyperref}
\usepackage{natbib}
\newcommand{\tabincell}[2]{\begin{tabular}{@{}#1@{}}#2\end{tabular}}  
\journal{Journal of \LaTeX\ Templates}

\begin{document}
%\begin{sloppypar}
\begin{frontmatter}

\title{Incorporating Surprisingly Popular Algorithm and Euclidean Distance-based Adaptive Topology into PSO}

\author[mymainaddress]{Xuan Wu \corref{first}}
\author[mymainaddress]{Jizong Han \corref{first}}
\author[mysecondaryaddress]{Di Wang}
\author[mythirdaddress]{Pengyue Gao}
\author[myfourthdaddress]{Quanlong Cui}
\author[myfifthaddress]{Liang Chen}
\author[mysixthaddress]{Yanchun Liang}
\author[mysevenththaddress]{Han Huang}
\author[myeighthaddress]{Heow Pueh Lee}
\author[mysecondaryaddress,myninthaddress]{Chunyan Miao}
\author[mymainaddress,mytenthaddress]{You Zhou\corref{mycorrespondingauthor}}
\author[mymainaddress]{Chunguo Wu\corref{mycorrespondingauthor}}
\cortext[first]{The first two authors contribute equally}
\cortext[mycorrespondingauthor]{Corresponding author}

\address[mymainaddress]{Key Laboratory of Symbolic Computation and Knowledge Engineering of Ministry of Education, College of Computer Science and Technology, Jilin University, Changchun, 130012, China}
\address[mysecondaryaddress]{Joint NTU-UBC Research Centre of Excellence in Active Living for the Elderly, Nanyang Technological University, 639798, Singapore}
\address[mythirdaddress]{State Key Laboratory of Superhard Materials and International Center for Computational Method and Software, College of Physics, Jilin University, Changchun 130012, China}
\address[myfourthdaddress]{Commercial Quality and Efficiency Department, Baidu Inc, Beijing, 100085, China}
\address[myfifthaddress]{Department of Computer Science, Shantou University, Shantou, 515063, China}
\address[mysixthaddress]{School of Computer Science, Zhuhai College of Science and Technology, Zhuhai, 519041, China}
\address[mysevenththaddress]{School of Software Engineering, the South China University of Technology, Guangzhou, 510641, China}
\address[myeighthaddress]{Department of Mechanical Engineering, National University of Singapore, 117575, Singapore}
\address[myninthaddress]{School of Computer Science and Engineering, Nanyang Technological University, 639798, Singapore}
\address[mytenthaddress]{College of Software, Jilin University, Changchun, 130012, China}

\begin{abstract}
While many Particle Swarm Optimization (PSO) algorithms only use fitness to assess the performance of particles, in this work, we adopt Surprisingly Popular Algorithm (SPA) as a complementary metric in addition to fitness. Consequently, particles that are not widely known also have the opportunity to be selected as the learning exemplars. In addition, we propose a Euclidean distance-based adaptive topology to cooperate with SPA, where each particle only connects to $k$ number of particles with the shortest Euclidean distance during each iteration. We also introduce the adaptive topology into heterogeneous populations to better solve large-scale problems. Specifically, the exploration sub-population better preserves the diversity of the population while the exploitation sub-population achieves fast convergence. Therefore, large-scale problems can be solved in a collaborative manner to elevate the overall performance. To evaluate the performance of our method, we conduct extensive experiments on various optimization problems, including three benchmark suites and two real-world optimization problems. The results demonstrate that our Euclidean distance-based adaptive topology outperforms the other widely adopted topologies and further suggest that our method performs significantly better than state-of-the-art PSO variants on small, medium, and large-scale problems. 

\end{abstract}


\end{frontmatter}

\section{Introduction}
\label{sec1}

%Surprisingly Popular Algorithm (SPA) is a highly effective crowd decision model, which well emphasizes the key knowledge mastered by the minority in the crowd. To better avoid Particle Swarm Optimization (PSO) falling into the local minima, in this research, we adopt SPA as the complementary metric other than fitness to assess the performance of particles in PSO. In addition, we propose an Euclidean distance based adaptive topology, where each particle only connects to $k$ number of particles with the shortest Euclidean distance during each iteration. To strive for a better balance between exploration and exploitation, we divide the population into two sub-populations, where one sub-population is expected to explore unknown solution regions and the other is expected to fine-search in the currently optimal solution regions. We name the overall algorithm Surprisingly Popular Algorithm-based Adaptive Euclidean Distance Topology Learning Particle Swarm Optimization (SpadePSO). To evaluate the performance of SpadePSO, we conduct extensive experiments on various optimization problems including the full CEC2014 benchmark suite and two real-world optimization problems. The experimental results show that our proposed adaptive topology based on Euclidean distance outperforms the other widely adopted topologies. Moreover, the experimental results suggest that SpadePSO performs significantly better than state-of-the-art PSO variants.

Many swarm intelligence algorithms and their variants have been proposed to solve diverse types of research and practical problems, such as ordinary differential equations optimization \citep{usman_inferring_2020}, hyperparameters optimization \citep{oyelade_ebola_2022}, neural architecture search \citep{junior_particle_2019,suganuma_evolution_2020}, engineering design \citep{abualigah_improved_2021, wang_committee-based_2017}, etc. Owing to its straightforward principle, usage of few parameters, and fast convergence rate, Particle Swarm Optimization (PSO) \citep{eberhart_new_1995} has been recognized as a popular method for solving single-objective \citep{bonyadi_particle_2017} and multi-objective optimization problems \citep{liu_coevolutionary_2019, steenkamp_scalability_2021}. 

To better avoid falling into trapping regions, many PSO variants aim to improve on the parameter setting, learning strategy, hybridization with other algorithms, and neighborhood topology \citep{xu_particle_2019}. In most PSO variants \citep{shi_modified_1998, kiran_recombination-based_2013, yang_level-based_2018}, the selection of the learning exemplars only depends on the particle's fitness. However, if not designed carefully, as a measuring metric, the fitness might only lead the population to a local trapping region, especially when the underlying problem is complex and multi-objective. Therefore, many multi-metric methods have been used to yield better results \citep{Laith_Feature_2019, George_Multi_2015,Alok_Feature_2014}. In \citep{xia_triple_2020}, both fitness and improvement rate of fitness are regarded as the key metrics to assess the performance of particles, because particles with a high improvement rate have a high probability of finding a better solution region. 
%By adopting this method, useful knowledge is used to better avoid PSO falling into trapping regions.

In fact, the selection method of the learning exemplars based on highly fit individuals implicitly adopts the idea of democratic voting, which is characterized by the majority advantage and the independence of individual judgment \citep{lorenz_how_2011}. Democratic voting, however, tends to emphasize the most popular opinion, not necessarily the most correct one. In fact, the most popular opinion may often be superficial and erroneous, while the correct answer may be not widely known and shared \citep{simmons_intuitive_2011,chen_eliminating_2004}. To overcome the limitations of democratic voting, Prelec et al. \cite{prelec_solution_2017} proposed a sociological decision-making method called Surprisingly Popular Algorithm (SPA) (also known as surprisingly popular decision), which could preserve the valuable knowledge of the minority. 

We introduced SPA as a complementary metric in addition to fitness in our previous work, named SPA-CatlePSO \citep{cui_surprisingly_2019}. If certain particles with high fitness values are connected by only a few particles in the topology, the other particles may not learn from those highly fit ones. Adopting the novel metric of SPA allows particles with high fitness values to share their own position with other particles. Instead of improving PSO through modifications to the parameter setting, learning strategy, hybridization with other algorithms, or neighborhood topology, we used SPA to evaluate the performance of particles in \citep{cui_surprisingly_2019}. However, owing to the loss of individual information of particles, e.g., position and velocity, in constructing the topology, SPA-CatlePSO may fail to provide insightful directional guidance for certain complex optimizing problems (see Section \ref{sec4}).

Considering the inefficiency of the particle topology used in SPA-CatlePSO, we propose an adaptive Euclidean distance-based topology based on the small-world network \citep{watts_collective_1998} to cooperate with SPA for performance improvement. In a small-world network, the probability of having a connection between two nodes with shorter Euclidean distance is correspondingly higher and vice versa (see Section \ref{sec2.1} for more details). By adopting this topology, the influence of individual particles is restricted to a local region, thus maintaining the diversity of the population \cite{newman_renormalization_1999}. To preserve the connectivity of the small-world network in our proposed topology, we stipulate that Particle $i$ connects to particles closer to it in the Euclidean space. In addition, to deal with the case where there is no excellent learning exemplar in the neighborhood of certain particles, Particle $i$ also connects to particles with high fitness values with the correspondingly higher probability. Furthermore, to better solve large-scale problems, the heterogeneity of particles is also ensured by dividing the population into exploration and exploitation sub-populations.  The exploration sub-population is expected to explore unknown solution regions while the exploitation sub-population is expected to fine-search in the currently optimal solution regions. In addition, particles in the same sub-population are closer to each other, while particles in different sub-population are farther apart. Our proposed approach is called Surprisingly Popular Algorithm-based Adaptive Euclidean Distance-based Topology Learning Particle Swarm Optimization (SpadePSO).

To assess the performance of SpadePSO, we conduct extensive experiments upon a series of optimization problems, including the full CEC2014 benchmark suite \citep{liang_problem_nodate}, the CEC2013 large-scale benchmark suite \citep{li2013benchmark}, the CEC2018 dynamic multi-objective optimization benchmark suite \citep{jiang2018benchmark}, the Spread Spectrum Radar Polyphase (SSRP) code design problem \citep{dukic_method_1990}, and the HIV model inference problem \citep{tian_latinpso_2019}. The experimental results on the full CEC2014 benchmark suite show that SpadePSO performs significantly better than PSO \citep{eberhart_new_1995}, TSLPSO \citep{xu_particle_2019}, HCLPSO \citep{lynn_heterogeneous_2015}, OLPSO \citep{zhan_orthogonal_2011}, GL-PSO \citep{gong_genetic_2016}, XPSO \citep{xia_expanded_2020}, and DMO \citep{agushaka_dwarf_2022} measured by the Wilcoxon signed ranks test. The experimental results on the CEC2013 large-scale benchmark suite demonstrate the superior performance of SpadePSO on higher dimensionality. On the SSRP code design and the HIV model inference problems, SpadePSO performs significantly better than all the compared PSO variants as well, in terms of fitness values, as validated by the one-tailed $t$-test.

The key contributions of this work are as follows:

\begin{itemize} 

\item
To cooperate with SPA, we propose an adaptive Euclidean distance-based topology inspired by small-world network connectivity and demonstrate its effectiveness in achieving better performance by conducting extensive experiments. 
\item
We propose an algorithm involving two sub-populations with an adaptive Euclidean distance-based topology. One sub-population better preserves the diversity of the population while the other achieves fast convergence. 
\item
We evaluate the performance of SpadePSO using three benchmark suites, and two real-world optimization problems. The experimental results indicate that SpadePSO performs significantly better than the conventional and state-of-the-art PSO variants.
\end{itemize} 

The remainder of this paper is organized as follows: Section \ref{sec2} introduces the related PSO variants and SPA. Section \ref{sec3} describes the adaptive Euclidean distance-based topology where each particle is connected to the neighboring particles. It also presents SpadePSO with two sub-populations. Section \ref{sec4} discusses the experimental results. Finally, Section \ref{sec5} draws the conclusion and proposes future work.

\section{Related Work}
\label{sec2}
In this section, we first introduce relevant PSO variants and then introduce SPA, its applications, and how to model SPA in PSO.

\subsection{PSO and its variants}
\label{sec2.1}
In the classical PSO \citep{eberhart_new_1995}, Particle $i$ is associated with two attributes, namely velocity $\bm{v}$ and position $\bm{x}$, whose update formulas are as follows: 
\begin{equation}
v_{i, j}=v_{i, j}+ {c_{1}} {r_{1,j}}\left(x_{i, j}^{pbest}-x_{i, j}\right) + {c_{2}} {r_{2,j}}\left(x_{ j}^{gbest}-x_{i, j}\right),
\end{equation}
\begin{equation}
x_{i, j}= x_{i, j} +v_{i, j}, 
\label{e2}
\end{equation}
\noindent where for the $j$th dimension, $v_{i, j}$ and $x_{i, j}$ respectively denote the velocity and position of Particle $i$, $r_{1, j}$ and $r_{2, j}$ are two uniformly distributed random numbers independently generated within the [0, 1] range, $x_{i,j}^{pbest}$ denotes the historical best position of Particle $i$, $x_j^{gbest}$ denotes the historical best position in the population, and $c_1$ and  $c_2$ are the acceleration coefficients.

To achieve a better balance between exploration and exploitation, four aspects, namely, the parameter setting, learning strategy, hybridization with other algorithms, and neighborhood topology, can be tweaked \citep{xu_particle_2019}. 

Parameter setting can control the convergence tendency of PSO to achieve a better balance between exploration and exploitation. Classical parameter settings such as the inertia weight \citep{shi_modified_1998} and the constriction coefficient \citep{clerc_particle_2002} can be set by conducting experiments. Instead of setting parameters through experiments, recent studies adjusted parameter settings adaptively by assessing the state of the population. For instance, Zhan et al. \cite{zhi-hui_zhan_adaptive_2009} divided the population into convergence, exploitation, exploration, and jumping-out categories according to the evolutionary state defined by the Euclidean distance between particles and then adaptively adjusted the inertia weights and acceleration coefficients. Liu \cite{liu_order-2_2015} concluded that PSO is stable, if and only if the inertia weight $w$ and acceleration coefficients $c_1$ and $c_2$ satisfy the following condition:
\begin{equation} 
 (1-w)\mu^2 + (1+w)\sigma^2 <(1+w)^2(1-w),
\end{equation}
where $w \in (-1,1), \mu=1+w-(c_1 + c_2)/2$, and $\sigma^2 = ({c_1}^2 + {c_2}^2)/12$. If the real-world application only requires a satisfactory level of accuracy, a stable PSO often performs better than an unstable PSO \citep{liu_order-2_2015}. 

Many learning strategies have been proposed to construct the learning exemplars to replace $\bm{x}^{gbest}$ and $\bm{x}^{pbest}$. For instance, Liang et al. \cite{liang_comprehensive_2006} proposed the Comprehensive Learning PSO (CLPSO) based on the Comprehensive Learning Strategy (CLS). Specifically, each particle learns from its own $\bm{x}^{pbest}$ with probability $\eta$ and learns from others' $\bm{x}^{pbest}$ determined by the tournament selection with probability $(1-\eta)$, where $\eta$ denotes the learning probability. The velocity update formula of CLPSO is as follows: 
\begin{equation}v_{i, j}=w v_{i, j}+ {c_{1}} {r_{1,j}}\left(x_{i, j}^{cl}-x_{i, j}\right), \label{e4}\end{equation}
where $x_{i,j}^{c l}$ denotes the learning exemplar constructed by CLS for the $j$th dimension of Particle $i$. By making the particle learn from others on each dimension with probability $(1-\eta)$, CLS helps particles escape from the local optima, enabling wider exploration. Zhan et al. \cite{zhan_orthogonal_2011} proposed the Orthogonal Learning PSO (OLPSO), which is designed to search for the best combination of $\bm{x}^{gbest}$ and $\bm{x}^{pbest}$ to construct the learning exemplars. Inspired by OLPSO, Xu et al. \cite{xu_particle_2019} adopted $\bm{x}^{gbest}$ and the combination of $\bm{x}^{pbest}$ and $\bm{x}^{gbest}$ as the learning exemplars, independently.

Hybridization of PSO with other algorithms is another focus of prior studies. For instance, Kiran et al. \cite{kiran_recombination-based_2013} combined $\bm{x}^{gbest}$ and the best solution of Artificial Bee Colony (ABC) to generate a new exemplar, named TheBest. Subsequently, TheBest is given to the populations of PSO and ABC as $\bm{x}^{gbest}$ and neighboring food source for onlooker bees, respectively. Gong et al. \cite{gong_genetic_2016} combined Genetic Algorithm (GA) and PSO to construct the learning exemplars and proposed GL-PSO. Yang et al. \cite{yang_level-based_2018} introduced the idea of Differential Evolution (DE) into PSO and generated particles randomly as the learning exemplars to solve large-scale optimization problems.

\begin{figure}[!t]
\centerline{
 \includegraphics[width=0.8\textwidth]{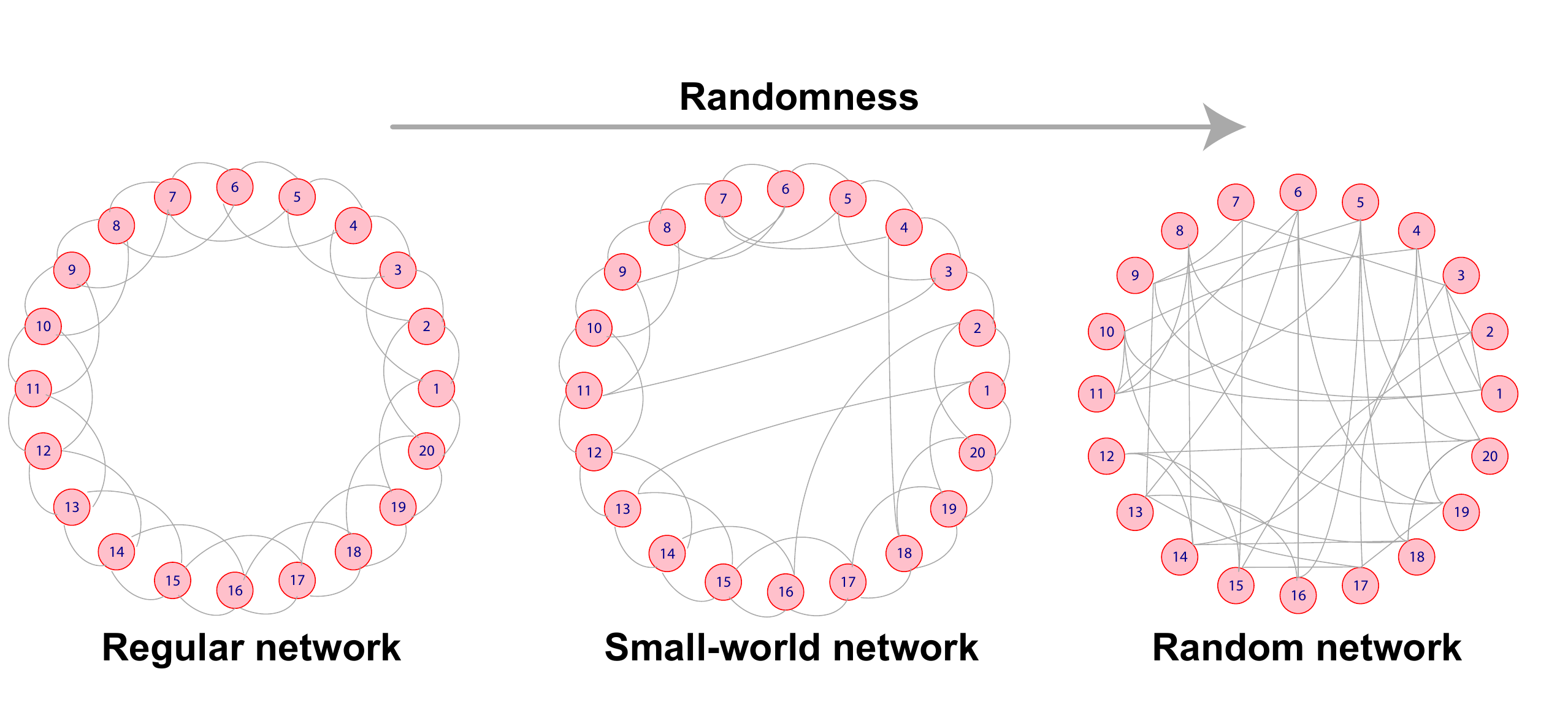}}
 \caption{An illustration of a small-world network \cite{newman_renormalization_1999}.}
\label{fig2}
\end{figure}

Depending on diverse information-sharing mechanisms, topology affects the balance between exploration and exploitation. As sub-populations can be assigned with different learning exemplars \citep{Zhao_Dynamic_2010,ghosh_inter-particle_2012,lynn_population_2018, wang_saturated_2019}, a large number of heterogeneous PSO variants were proposed. For instance, Lynn et al. \cite{lynn_heterogeneous_2015} proposed the Heterogeneous Comprehensive Learning PSO (HCLPSO), which comprises  the exploration as well as the exploitation sub-populations. The particles in the exploration sub-population only learn from $\bm{x}^{cl}$ in the same sub-population according to Eq. \eqref{e4}; whereas the particles in the exploitation sub-population learn from $\bm{x}^{cl}$ in the whole population and $\bm{x}^{gbest}$ as follows:
\begin{equation}
v_{i, j}=w v_{i, j}+c_{1} r_{1, j}\left(x_{i, j}^{cl}-x_{i, j}\right)+c_{2} r_{2, j}\left(x_{j}^{g b e s t}-x_{i, j}\right). \end{equation}
\noindent Yang et al. \cite{yang_level-based_2018} divided particles equally into four levels according to the descending order of their fitness values. Specifically, level $L_4$ learns from levels $L_{1,2,3}$, level $L_3$  learns from levels $L_{1,2}$, level $L_2$  learns from level $L_{1}$, and level $L_1$ remains constant. In addition, the small-world network \citep{watts_collective_1998} is also a popular topology of PSO. As shown in Figure~\ref{fig2}, a small-world network is constructed based on a regular ring topology network of $n$ vertices. In a regular ring topology, each vertex is connected to its $k$ nearest neighbors by undirected edges. To construct the small-world network, all edges of the regular ring topology may be rewired. Specifically, the vertex at one end remains unchanged, while the vertex at the other end is randomly selected with probability $p$. Given that a small-world network could suppress the influence of individual particles and maintain the diversity of the population \cite{newman_renormalization_1999}, it has been widely used for the connecting topologies in PSO \citep{gong_small-world_2013,qiu_novel_2020,liu_niching_2020}. Other than heterogeneous PSO variants and small-world networks, Xia et al. \cite{xia_expanded_2020} generated random permuted order numbers for particles, and assign each particle's left and right neighbors according to the permuted order number.

% Waintraub et al. \cite{waintraub2007cellular} arranged particles into a two-dimensional cellular topology as shown in Figure~\ref{fig3}.
%\begin{figure}[!t]
%\centerline{
 %\includegraphics[width=0.5\textwidth]{figs/3.pdf}}
%\caption{An illustration of a cellular topology with $m$ rows and $n$ columns \cite{waintraub2007cellular}.}

%\label{fig3}
%\end{figure}

Including heterogeneous PSO \citep{xu_particle_2019, lynn_heterogeneous_2015}, the topologies of many PSO variants are constructed based on regular networks, which may make them easily fall into local minima \citep{Mendes_The_2004}. To counter this, we propose a Euclidean distance-based topology based on small-world networks and introduce it into the heterogeneous PSO. In addition, in most PSO variants, the choice of the learning exemplars can be divided into four categories, $\bm{x}^{gbest}$, its own $\bm{x}^{pbest}$, $\bm{x}^{pbest}$ of particles with high fitness values, and random particles. The first three learning exemplars are selected merely based on the evaluation of their fitness values, while the randomly selected learning exemplars have no theoretical basis and are often ineffective \citep{xu_particle_2019}. In this work, we use the surprisingly popular degree as an additional metric to evaluate the performance of particles beyond the mere evaluation of fitness.

\subsection{SPA and its applications}
\label{sec2.2}

\begin{figure}[!t]
\centerline{
 \includegraphics[width=1\textwidth]{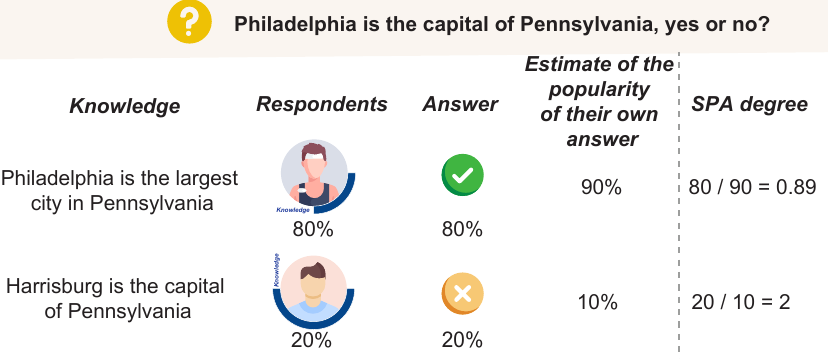}}
 \caption{An illustration of SPA. Each people gives the answer based on what he/she thinks of and an estimate of the popularity of the answer. By computation, the importance of the correct answer can be leveraged.}
\label{fig1}
\end{figure}   

To preserve the potentially correct knowledge that may not be widely known among the overall population, SPA hinges on asking people two questions, namely what they think the right answer is, and how popular they think the answer will be. For example, as shown in Figure~\ref{fig1}, when the question of whether Philadelphia is the capital of Pennsylvania is asked, 80$\%$ of the respondents may only recall that Philadelphia is a large, famous, historically significant city in Pennsylvania, and hence conclude mistakenly that it is the capital of Pennsylvania. The remaining 20$\%$ who vote ``no'' possess the correct knowledge that the capital of Pennsylvania is Harrisburg. As illustrated in Figure~\ref{fig1}, people with different knowledge have different perceptions of the popularity of their answers. People who know that Harrisburg is the capital of Pennsylvania expect the popularity of their answer to be low. However, the rest of the respondents believe that most respondents have the same answer as they do. By computing the surprisingly popular degree, i.e., the ratio of the actual turnout over the estimate for the popularity of a given answer, the correct answer ``no'' is leveraged. In this and many other cases, the answer having a higher surprisingly popular degree is the correct answer. To better capture the integrity of the knowledge via the perceived popularity of the corresponding knowledge, based on the inquiries of popularity, SPA leverages the importance of the correct answer by assigning high voting weights to more confident answers \citep{prelec_solution_2017, hosseini_surprisingly_2021}.

As SPA \citep{prelec_solution_2017} can identify the knowledge possessed by the minority by asking people two questions, it has been widely used in social science, computer science, and other disciplines. For instance, Lee et al. \cite{lee_testing_2018} used SPA to make more accurate predictions of the winners of National Football League (NFL) games and found that SPA could predict better than many NFL media. To solve classification problems, Luo et al. \cite{luo_machine_2022} asked each classifier to predict the performance of the other classifiers and learn the feedback of the other classifiers. As mentioned earlier, the original study proposing SPA \citep{prelec_solution_2017} focused on choosing the right answer from a list of the alternatives. Hosseini et al. \cite{hosseini_surprisingly_2021} extended SPA to give a ground-truth rank of the alternatives. Cui et al.  \cite{cui_surprisingly_2019} introduced SPA into PSO to construct the learning exemplars $\bm{x}^{sbest}$ to replace $\bm{x}^{gbest}$.
%To better understand how to model SPA in PSO and carry out the follow-up work, we give the following three concepts of the SPA calculation process. $Knowledge \quad Prevalence \quad Degree$: It is the proportion of the total number of particles in the population that has the privilege to access the fitness of a particle. $Expected \quad Turnout$: It is the population's estimation for the popularity of a given learning exemplar, which is the sum of each particle's estimation. $Surprisingly \quad Popular \quad Degree$: It is the ratio of the actual turnout to the expected turnout.

\begin{table}[!t]
	\centering
	\caption{Notations used in this paper}
	\label{table1}
	%\begin{adjustbox}{center}
	\scalebox{0.65}{
	\begin{tabular}{llll}
		\toprule
		Notation & Definition  \\
		\midrule
		$G\left(\bm{V}, \bm{E}\right)$ & Graph $G$ with vertex set $\bm{V}$ and edge set $\bm{E}$  \\
		$G\left(\bm{V}, \bm{E}_{exp}\right)$ & Graph $G$ with vertex set $\bm{V}$ and edge set $\bm{E}_{exp}$  \\
		$G\left(\bm{V}, \bm{E}_{tmp}\right)$ & Graph $G$ with vertex set $\bm{V}$ and edge set $\bm{E}_{tmp}$  \\
		$\bm{E}$ & Edge set corresponding to the distance information\\
		$\bm{E}_{exp}$ & \tabincell{l}{Edge set randomly generated by particles with high fitness values}   \\
		$\bm{E}_{tmp}$ & Edge set generated by $\bm{E}\cup\bm{E}_{exp}$\\
		$\bm{A}$ & Adjacency matrix	corresponding to $G\left(\bm{V}, \bm{E}\right)$   \\
		$\bm{A_{exp}}$ & Adjacency matrix corresponding to $G\left(\bm{V}, \bm{E}_{exp}\right)$   \\
		$\bm{A_{tmp}}$ & Adjacency matrix corresponding to $G\left(\bm{V}, \bm{E}_{tmp}\right)$   \\
		$f(\cdot)$ & Fitness function   \\
		$at$ & Actual turnout \\
		$et$ & Expected turnout   \\
		$kp$ & Knowledge prevalence degree   \\
		$\theta$ & Surprisingly popular degree   \\
		$n_{exp}$ & Number of experts   \\
		$d_{i,j}$ & Euclidean distance between particles $i$ and $j$ \\
		$\bm{T}(n_{exp})$ &  Set of the top $n_{exp}$ particles according to the descending order of fitness values  \\
		${i}_{j}$ & Descending ranking order of the particle $j$ according to its fitness value \\
		${p}\left(i_{j}\right)$ & Connection probability in $\bm{E}_{exp}$ \\
		$k$ & Out-degree of each particle in $G\left(\bm{V}, \bm{E}\right)$  \\
		$v_k$ & Increasing velocity of  $k$ \\
		$k_0$ & \tabincell{l}{Out-degree of each particle in $G\left(\bm{V}, \bm{E}\right)$ in the initial iteration} \\
		\bottomrule
	\end{tabular}}
	%\end{adjustbox}
\end{table}

\begin{figure}[!t]
	\centering
	\includegraphics[scale=0.34]{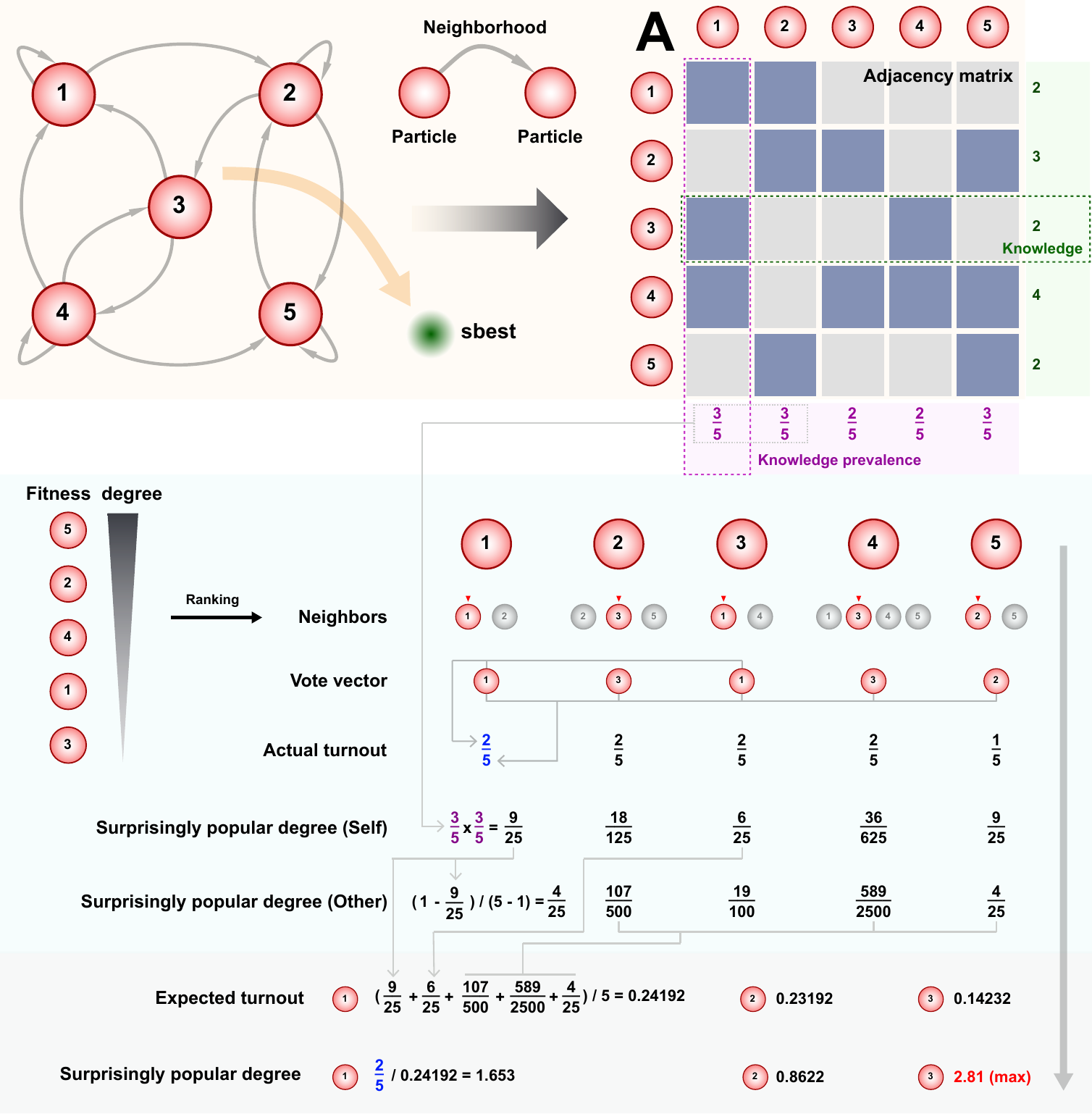}
	\caption{ An illustration on the computation of the surprisingly popular degree \citep{cui_surprisingly_2019}.}
	\label{fig4}
\end{figure}

As a fundamental work \citep{cui_surprisingly_2019}, it is necessary for us to review the technical details (all notions used are presented in Table~\ref{table1}). Let $\bm{V}$ and $\bm{E}$ denote the vertex set and the edge set of PSO with $n$ particles, respectively. The directed edge $e_{i,j} \in \bm{E}$ means that Particle $i$ knows the fitness value of Particle $j$, and is able to learn from it, where $i$, $j \in \bm{V}$. Thus, PSO can be represented by a directed graph $G\left(\bm{V}, \bm{E}\right)$, named the knowledge transfer topology. The asymmetry adjacency matrix $\bm{A}=\left(a_{i,j}\right)_{n \times n}$ of $G\left(\bm{V}, \bm{E}\right)$ is defined as follows:

\begin{small}
\begin{equation}
	a_{i,j}=\left\{\begin{array}{ll}
		1, & e_{i,j} \in E, \\
		0, & e_{i,j} \notin E,
	\end{array} \quad i,j \in \{1,2, \cdots, n\}\right. ,
\label{e6}
\end{equation}
\end{small}
where $i$ and $j$ may be set to the same index value.

Figure~\ref{fig4} illustrates the knowledge transfer topology and the asymmetry adjacency matrix $\bm{A}$ using an illustrating PSO population of five particles. The surprisingly popular degree is computed as follows: first, Particle $i$ selects the particle with the maximum fitness value from its first-order neighbors given by the knowledge transfer topology $G\left(\bm{V}, \bm{E}\right)$, named the expected learning exemplar $\bm{x}_{j_{i}^{*}}$. The index $j_{i}^{*}$ is determined as follows:
\begin{equation}j_{i}^{*}=\arg \max _{a_{i j}=1}\left\{f\left(\bm{x}_{j}\right)\right\}, i,j \in \{1,2, \cdots, n\},\end{equation}
where $f(\cdot)$ denotes the fitness function. The expected learning exemplars of all particles $J^{*}$ are defined as follows:
\begin{equation}
J^{*}=\left(j_{1}^{*}, j_{2}^{*}, \cdots, j_{n}^{*}\right).
\end{equation}
For example, $J^{*}=(1,3,1,3,2)$ in Figure~\ref{fig4}.

Multiple particles may select the same particle as the expected learning exemplar, such as Particle 1 and Particle 3 ($j^*_1 = j^*_3 = 1$) as shown in Figure~\ref{fig4}. Therefore, it is necessary to count the total number of unique exemplars. Let $\boldsymbol{C}$ denote the set of the expected learning exemplars, where $\bm{C}= unique(J^*)$. Let $M=|{\bm{C}}|$, so there are $M$ number of unique particles being selected as the expected learning exemplars. As shown in Figure~\ref{fig4}, $\bm{C}=\{1,3,2\}$ and $ M=3$.

Next, the actual turnout $at_{j}$ is defined as the ratio of the votes of Particle $j$ over the population size, which is computed as follows: 
\begin{equation}
at_{j}=\left\{\begin{array}{cc}
		\left(\sum_{i=1}^{n} a_{i j}\right) / n, & j \in C, \\
		0, & \text{ otherwise. }
	\end{array}\right.	
\end{equation}
%\left(\sum_{i=1, j_{i}^{*}=j}^{n} a_{i j}\right) / n, & j \in C, \\
As shown in Figure~\ref{fig4}, $at_{1} = 2/5$, $at_{2} = 1/5$, and $at_{3} = 2/5$.

To derive the surprisingly popular degree, each particle has to give the expected learning exemplar and the expected turnout of all the expected learning exemplars. Then, the expected turnout is computed subsequently. For Particle $i$, the expected turnout of all the expected learning exemplars can be divided into two categories: the expected turnout of its own expected learning exemplar and the turnout of the other expected learning exemplars. To obtain the expected turnout, the knowledge prevalence degree $kp_{k}$ of Particle $k$ is computed as follows: 
\begin{equation}
kp_{k}=\left(\sum\nolimits_{i=1}^{n} a_{i k}\right) / n , k \in \{1,2, \cdots, n\}.
\end{equation}
As shown in Figure~\ref{fig4}, $kp_{1} = 3/5$, $kp_{2} = 3/5$, $kp_{3} = 2/5$, $kp_{4} = 2/5$ and $kp_{5} = 3/5$.

Subsequently, for Particle $i$, the expected turnout of $j_{i}^{*}$ is computed as follows: 
\begin{equation}
\alpha_{i, j_{i}^{*}}=\prod\nolimits_{\left\{k | k \in \{1,2, \cdots, n\}, a_{i k}=1\right\}} kp_{k}.
\end{equation}
As shown in Figure~\ref{fig4}, $\alpha_{1,1}=9/25$.

Furthermore, for Particle $i$, because all the expected learning exemplars except $\boldsymbol{x}_{j_{i}^{*}}$ share the same popularity, i.e., $1-\alpha_{i, j_{i}^{*}}$, their expected turnout is assumed to be equal and can be computed as follows:
\begin{equation}
\alpha_{i,j}=\left(1-\alpha_{i, j_{i}^{*}}\right)/(n-1), j \in \boldsymbol{C}, j \neq j_{i}^{*}.
\end{equation}
As shown in Figure~\ref{fig4}, $\alpha_{1,2}=\alpha_{1,3}=4/25$.

Finally, the averaged summarization of popularity from all particles is taken as the expected turnout $et_{j}$ of Particle ${j}$, defined as follows:
\begin{equation}
et_{j}=\left(\sum\nolimits_{i=1}^{n} \alpha_{i,j}\right) / n.
\end{equation}
As shown in Figure~\ref{fig4}, $et_{1}=0.24$, $et_{2}=0.23$, and $et_{3}=0.14$.

Hereby, the surprisingly popularity degree $\theta_{j}$ of Particle $j$ is defined as follows: 
\begin{equation}
\theta_{j}=at_{j} / et_{j}.
\end{equation}
As shown in Figure~\ref{fig4}, $\theta_{1}=1.65$, $\theta_{2}=0.86$, and $\theta_{3}=2.81$.

The learning exemplar $\boldsymbol{x}^{ {sbest}}$, which is the particle with the maximal surprisingly popular degree, is then identified as follows:
\begin{equation}k^{*}=\arg \max _{j \in \bm{C}}\left\{\theta_{j}\right\}, \end{equation}
\begin{equation}\bm{x}^{ {sbest}}=\bm{x}_{k^{*}}.
\label{e16}
\end{equation}

As shown in Figure~\ref{fig4}, Particle 3 has the maximal surprisingly popular degree of 2.81; therefore, the learning exemplar $\boldsymbol{x}^{ {sbest}}$ is identified as $\boldsymbol{x}_3$.

One of the most important components in SPA-CatlePSO \citep{cui_surprisingly_2019} is the knowledge transfer topology $G\left(\bm{V}, \bm{E}\right)$, where each particle selects the expected learning exemplar to guide the population searching direction. Therefore, topology has considerable inﬂuence on the performance of SPA-CatlePSO. The topology of SPA-CatlePSO is defined as follows. In the initial iteration, Particle $i$ unidirectionally connects to particles numbered from $i+1$ to $i+k$, where $k$ denotes a predefined number of edges for each particle. During each iteration, each particle is connected to particles with high fitness values with the correspondingly higher probability temporarily \citep{cui_surprisingly_2019}. In addition, if Particle $i$ selects $\bm{x}^{ {sbest}}$ as the learning exemplar and the fitness value of Particle $i$ is improved in the current iteration, Particle $i$ is connected to the particle corresponding to $\bm{x}^{ {sbest}}$ in the subsequent iteration. However, the topology proposed in \citep{cui_surprisingly_2019} cannot reflect the positional information of particles. Therefore, in this work, we propose a new knowledge transfer topology aiming for performance improvement by properly modeling the positional information.

%Specifically, SPA-CatlePSO takes the particle with the maximal surprisingly popular degree as the learning exemplar to guide the population searching direction and constructs an adaptive topology in PSO. To derive the surprisingly popular degree, particle $i$ is asked to answer the following two questions, namely which particle among particle $i$'s first-order neighbors is suitable to be used to guide the population searching direction, and how popular each particle will be. By answering these two questions, $\bm{x}^{ {sbest}}$ replaces the population's historical best position $\bm{x}^{gbest}$ as the learning exemplar to guide the population searching direction. The topology of SPA-CatlePSO \citep{cui_surprisingly_2019} is defined as follows. Particle $i$ unidirectionally connects to particles numbered from $i+1$ to $i+k$ in the initial iteration, where $k$ denotes the predefined number of neighbors (see Section \ref{sec2} for more details). Due to the loss of individual information of particles, e.g., position and velocity, in constructing the topology, it is found that SPA-CatlePSO tends to fail to provide insightful directional guidance for some certain complex  optimizing problems (see Section \ref{sec4}).
\section{SpadePSO}
\label{sec3}
In this section, we propose an adaptive Euclidean distance-based topology inspired by the small-world networks, introduce SpadePSO, and analyze the algorithm complexity and the diversity of sub-populations in SpadePSO. 

\subsection{Adaptive Euclidean distance-based topology}
\label{sec3.1}

Watts and Strogatz \cite{watts_collective_1998} showed that information transmission through social networks is affected by three characteristics of the topology structure: the number of clusters, the number of neighbors, and the average shortest path length between two nodes. To propose an appropriate topology, we need to consider the relative position information of particles in a multidimensional space, because the topology composed of particles with similar positions has excellent clustering performance \citep{watts_collective_1998}.

In this paper, we propose a new knowledge transfer topology $G\left(\bm{V}, \bm{E}\right)$. In the initial iteration, the edge set $\bm{E}$ is constructed based on the Euclidean distance information. Specifically, each particle connects to the first $k$ particles having the shortest Euclidean distance and can know the fitness values of these $k$ particles, i.e., the out-degree value \cite{su_efficient_2022} of each particle is $k$. The Euclidean distance is defined as follows:
\begin{equation}
d_{i,j} = \sqrt{\sum\nolimits_{l=1}^{D} {(x_{i,l}- x_{j,l})^2}}, 
\label{e17}
\end{equation}
where $d_{i,j}$ denotes the Euclidean distance between Particles $i$ and $j$. Note that for Particle $i$, the first $k$ particles include Particle $i$ itself, because the distance from Particle $i$ to itself is always 0, i.e., always the shortest. In addition, the in-degree value \cite{su_efficient_2022} of Particle $i$ represents how many particles can know its fitness value and is determined by the Euclidean distance between Particle $i$ and other particles. The closer Particle $i$ is to multiple particles, the higher its in-degree value, because it connects to more particles.

The process of small-world network construction involves the risk of breaking the network connectivity, generating isolated clusters, or accidentally deleting the key connection \citep{wanliang_wang_research_2013}. To mitigate these concerns, Newman and Watts \cite{newman_renormalization_1999} proposed the NW small-world network, where the number of edges increases rather than remains unchanged. To increase the number of edges, the out-degree of each particle ${k}$ in $G\left(\bm{V}, \bm{E}\right)$ increases linearly for the successive iteration, with the updating formula given as follows:  
\begin{equation}k=\left\lfloor k_0 + \left(v_k \cdot \frac{t}{T}\right)\right\rfloor, \label{e18}\end{equation}
where $k_0$ denotes the predefined out-degree value in the initial iteration, $v_k$ denotes a predefined parameter indicating the increasing velocity of out-degree $k$, $t$ and $T$ denote the current and the total iteration numbers, respectively, and $T$ is predefined.

\begin{figure}[!t]
	\centering
	\includegraphics[scale=0.7]{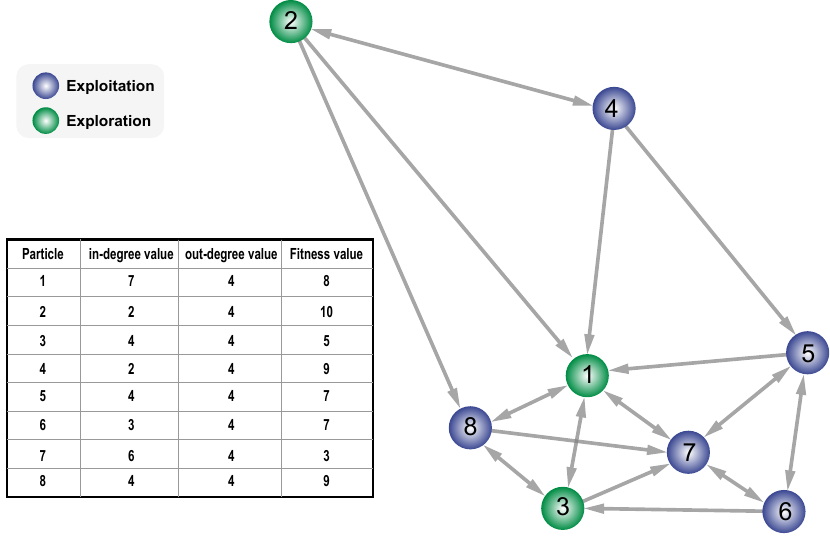}
	\caption{An illustration of connection relationships between particles. The population has 8 particles, and all particles have the same out-degree value, but have different in-degree values.}
	\label{fig5}
\end{figure}

As shown in Figure~\ref{fig5}, to further understand the definitions of out-degree and in-degree, we illustrate the connection relationships between particles. In Figure~\ref{fig5}, ${k}$ is set to four (the graph has self-loops wherein each particle has a self-connected edge not shown in the figure for brevity), the exploration sub-population size is set to three, the exploitation sub-population size is set to five, and the dimension is set to two (heterogeneous populations are described in the following subsection). Therefore, each particle connects to four neighboring particles (inclusive of itself) with the shortest Euclidean distance. As mentioned earlier, all particles have the same out-degree value but different in-degree values. As shown in Figure~\ref{fig5}, it is obvious that Particle 1, which resides in the center of the graph, has the highest in-degree value of seven as it is connected by Particles 1, 2, 3, 4, 5, 7, and 8. Particles 2 and 4, located at the boundary of the current search space, have the smallest in-degree value of 2 as they are only connected by each other.

Besides $G\left(\bm{V}, \bm{E}\right)$, we use the temporary directed graph $G\left(\bm{V}, \bm{E}_{exp}\right)$ to deal with the case where there is no excellent learning exemplar in the neighborhood of certain particles. During each iteration, $G\left(\bm{V}, \bm{E}_{exp}\right)$ is generated by $n_{exp}$ number of particles having the highest fitness values, called experts, with a certain probability ${p}\left(i_{j}\right)$ obtained as follows:

\begin{equation}{p}\left(i_{j}\right)=\left(\begin{array}{c}
		n-i_{j} \\
		n_{e x p^{-1}}
	\end{array}\right) \bigg/ \left(\begin{array}{c}
	n \\
	n_{exp}
	
\end{array}\right),i_{j} \in \bm{T}(n_{e x p}),
\label{e19}
\end{equation}
\noindent where $( \cdot )$ denotes the combinatorial function, ${i}_{j}$ denotes the descending ranking order, and $\bm{T}(n_{e x p})$ denotes the set of experts. The generating rule of the adjacency matrix $\bm{A}_{exp} = \left(b_{i, j}\right)_{n \times n}$ of $G\left(\bm{V}, \bm{E}_{exp}\right)$ is obtained as follows:

\begin{small}
\begin{equation}b_{i, j}=\left\{\begin{array}{ll}
		1, & \text { if } j \in T\left(n_{e x p}\right) \text { and } {rand()}<{p}\left(i_{j}\right), \\
		0, & \text { otherwise, }
	\end{array}\right. \label{e20}
\end{equation}
\end{small}

\noindent where $j \in \{1,2, \cdots, n\}$, and $rand()$ produces a uniformly distributed random number generated within the [0, 1] range. As shown in Figure~\ref{fig5}, there is no excellent learning exemplar in Particle 2's first-order neighbors. With the help of $G\left(\bm{V}, \bm{E}_{exp}\right)$, Particle 2 may connect to Particle 7. In each iteration, the joint-directed graph $G\left(\bm{V}, \bm{E}_{tmp}\right)$, i.e., $G_{t}\left(\bm{V}, \bm{E}\cup\bm{E}_{exp}\right)$, is used to construct the learning exemplars in SpadePSO. Specifically, the adjacency matrix $\bm{A}_{tmp} =\left(t_{i,j}\right)_{n \times n} $ of $G\left(\bm{V}, \bm{E}_{tmp}\right)$ is obtained as follows:
\begin{equation}t_{i,j}=\left\{\begin{array}{ll}
		1, & \text { if } a_{i, j} =1 \text{ or }  b_{i, j} =1, \\
		0, &   \text{ otherwise},
	\end{array}\right. \label{e21}
\end{equation}
\noindent where $i,j \in \{1,2, \cdots, n\}$. Through this construction of $\bm{A}_{tmp}$, all particles can identify their excellent learning exemplars, which are generated according to either $\bm{A}$ or $\bm{A}_{exp}$. When $a_{i, j} =1$, $\bm{A}$ is used, and vice versa.

\begin{figure}[!t]
	\centering
	\includegraphics[scale=0.18]{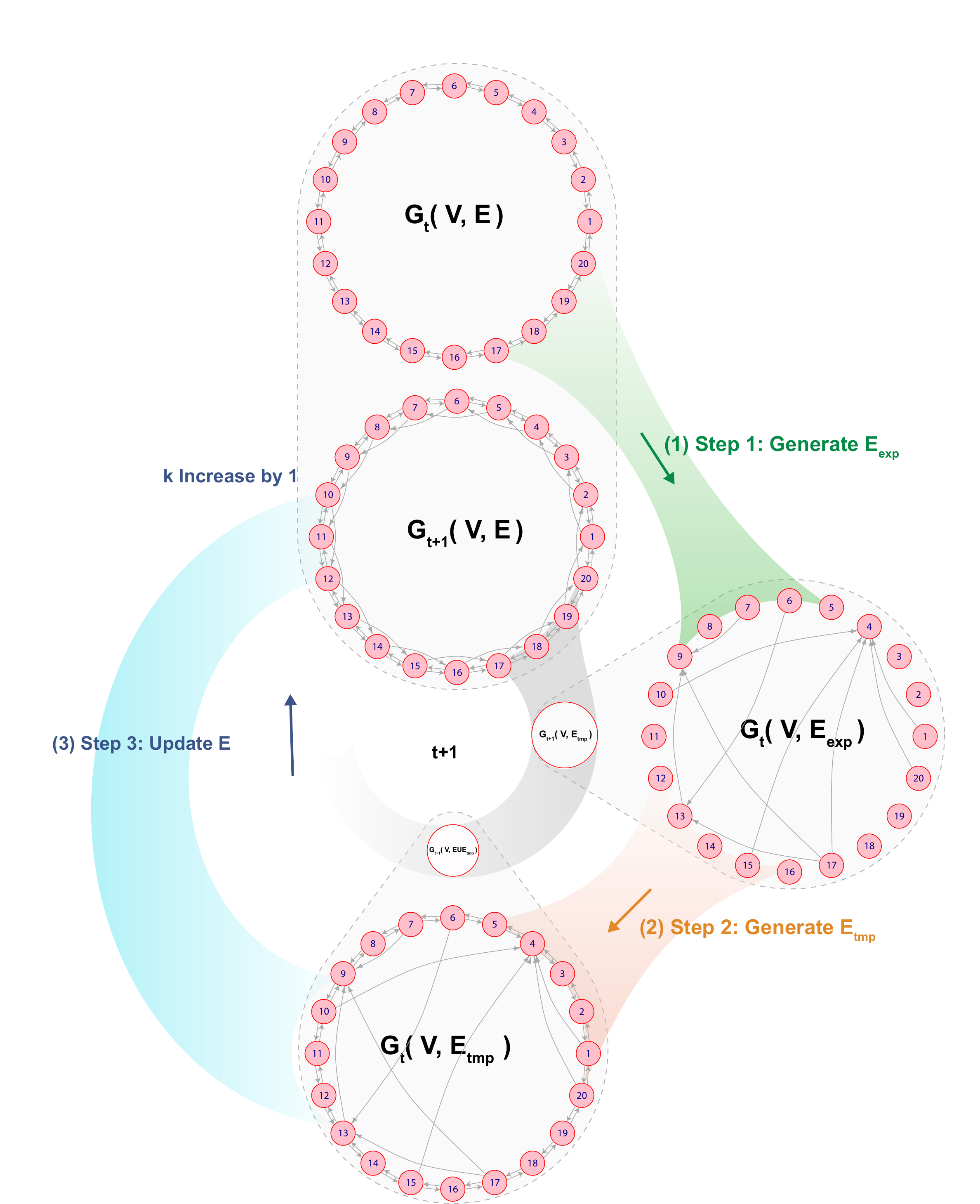}
	\caption{An illustration of the update process of $G_{t}\left(\bm{V}, \bm{E}\right)$ and the generation process of $G_{t}\left(\bm{V}, \bm{E}_{exp}\right)$ and $G_{t}\left(\bm{V}, \bm{E}_{tmp}\right)$.}
	\label{fig6}
\end{figure}

As shown in Figure~\ref{fig6}, we illustrate the update process of $G_{t}\left(\bm{V}, \bm{E}\right)$ and the generation process of $G_{t}\left(\bm{V}, \bm{E}_{exp}\right)$ and $G_{t}\left(\bm{V}, \bm{E}_{tmp}\right)$. To ensure that each particle is always connected to particles having the shortest Euclidean distance, we update $G_{t}\left(\bm{V}, \bm{E}\right)$ during each iteration. In addition, as mentioned earlier, $G_{t}\left(\bm{V}, \bm{E}_{exp}\right)$ and $G_{t}\left(\bm{V}, \bm{E}_{tmp}\right)$ are generated to avoid the situation where there is no excellent learning exemplar in the neighborhood of certain particles. Assuming ${k}$ in $G_{t}\left(\bm{V},\bm{E}\right)$ is set to 3 in the $t$th iteration,  Particle $i$ connects to two particles with the shortest Euclidean distance and connects to itself. Assuming ${k}$ increases by 1 according to Eq. \eqref{e19}, Particle $i$ connects to three particles with the shortest Euclidean distance and connects to itself for the successive iteration, as shown in the central part of Figure~\ref{fig6}.

\subsection{Heterogeneous populations}
\label{sec3.2}

To better solve large-scale problems, we propose SpadePSO, which comprises two heterogeneous sub-populations. The velocity of particles in the exploration sub-population is updated according to Eq. \eqref{e4}. Besides $\bm{x}^{cl}$, $\bm{x}^{sbest}$ is also regarded as the learning exemplar in the exploitation sub-population. The velocity update formula of particles in the exploitation sub-population is then given as follows: 

\begin{small}
\begin{equation}v_{i, j}=w v_{i, j}+c_{1} r_{1, j}\left(x_{i, j}^{c l}-x_{i, j}\right)+c_{2} r_{2, j}\left(x_{j}^{ {sbest}}-x_{i, j}\right), \label{e22}
\end{equation}
\end{small}

\noindent where $x_{j}^{sbest}$ denotes the learning exemplar constructed by SPA on the $j$th dimension, which is used to guide the exploitation sub-population searching direction. 

\begin{algorithm}[!t]  
  \caption{SpadePSO algorithm}  
  \label{spadepso}  
  \begin{algorithmic}[1]
  \State Randomly initialize $\bm{x}$ and $\bm{v}$ of all particles
  \State Evaluate the fitness value of $\bm{x}$
  \State Construct learning exemplar $\bm{x}^{cl}$ according to CLS 
  \State Construct the knowledge transfer topology $G \left(\bm{V},\bm{E}\right)$ (see Eq. \eqref{e17})
  \State Construct learning exemplar $\bm{x}^{sbest}$ according to $G \left(\bm{V},\bm{E}\right)$ (see Eqs. \eqref{e6}$\sim$ \eqref{e16}) 
  \While {(\textit{FEs} $\leq $ \textit{Max\_FEs})}
  		\State Update $\bm{v}$ of particles in the exploration sub-population (see Eq. \eqref{e4})
  		\State Update $\bm{v}$ of particles in the exploitation sub-population (see Eq. \eqref{e22})
  		\State Update $\bm{x}$ of all particles (see Eq. \eqref{e2})
  		\State Update the learning exemplar $\bm{x}^{cl}$ according to CLS
 		\State Update the knowledge transfer topology $G \left(\bm{V},\bm{E}\right)$ (see Eq. \eqref{e17})
 		\State Construct graph $G \left(\bm{V},\bm{E}_{exp}\right)$ (see Eq. \eqref{e20})
 		\State Unite $G \left(\bm{V},\bm{E}\right)$ and $G \left(\bm{V},\bm{E}_{tmp}\right)$ to construct $G \left(\bm{V},\bm{E}_{tmp}\right)$ (see Eq. \eqref{e21})
  		\State Construct the learning exemplar $\bm{x}^{sbest}$ according to $G \left(\bm{V},\bm{E}_{tmp}\right)$ (see Eqs. \eqref{e6} $\sim$ \eqref{e16})
  \EndWhile
  \end{algorithmic}  
\end{algorithm}

In the exploration sub-population, particles learn only from $x^{pbest}$ of its own and the other particles in the same sub-population. Hence, there is no accumulation of population common-learning experience regarding the searching direction in the exploration sub-population. Therefore, the exploration sub-population generally has a wider exploration ability and high population diversity \cite{xu_particle_2019, lynn_heterogeneous_2015}. As $G\left(\bm{V}, \bm{E}_{tmp}\right)$ covers the entire population and $\bm{x}^{sbest}$ is determined by the entire population, particles in the exploitation sub-population can learn from the best experience among the entire population. Therefore, the exploitation sub-population generally has a finer exploitation ability \cite{xu_particle_2019, lynn_heterogeneous_2015}. The pseudocode of SpadePSO is presented in Algorithm \ref{spadepso}, where \textit{FEs} and \textit{Max\_FEs} denote the current and total number of fitness function evaluations predefined, respectively. The source code of SpadePSO is available online\footnote{URL: https://github.com/wuuu110/SpadePSO}.

\subsection{Algorithm complexity analysis}
\label{sec3.3}

For PSO, the computation cost is determined by the initialization, evaluation, velocity update, and position update costs, each of which has a complexity of $O(nD)$, where $n$ and $D$ denote the population size and dimension, respectively. Compared with PSO, SpadePSO needs to additionally construct the learning exemplars $\bm{x}^{cl}$ and $\bm{x}^{sbest}$. For each particle in the exploration and exploitation sub-populations, its own $\bm{x}^{cl}$ is constructed. Therefore, the complexity of constructing $\bm{x}^{cl}$ is $O(n_1D)+O(n_2D)$, where $n_1$ and $n_2$ denote the exploration and exploitation sub-populations size, respectively. Conversely, only particles in the exploitation sub-population need to construct $\bm{x}^{sbest}$. To construct $\bm{x}^{sbest}$, a series of formulae having with the complexity $O(n_2)$ are used (see Eqs. \eqref{e6}$\sim$ \eqref{e16}). Therefore, the overall complexity of SpadePSO is $O(nD)$, which is the same as that of PSO.
\subsection{Population diversity in SpadePSO}
\label{sec3.4}

Population diversity can be regarded as a metric to measure the balance between exploration and exploitation \citep{olorunda_measuring_2008}. In this subsection, we compare the diversities of the exploration sub-population, the exploitation sub-population, and the entire population of SpadePSO. The population diversity $Div$ is computed as follows:

\begin{equation}
Div=\bigg{(}\sum\nolimits_{i=1}^{n} \sqrt{\sum\nolimits_{j=1}^{D} {(x_{i,j}- \overline{x}_{j})}^2 } \bigg{)} / n, 
\label{e23}
\end{equation}

\begin{equation}
\overline{x}_{j}=(\sum\nolimits_{i=1}^{n} {x_{i,j}}) / n, 
\label{e24}
\end{equation}
where $\overline{x}_{j}$ denotes the center position of the population on the $j$th dimension.

\begin{figure}[!t]
	\centering
	\subfigure[F1 function]{
	\begin{minipage}[t]{0.45\linewidth}
	
	\includegraphics[scale=0.4]{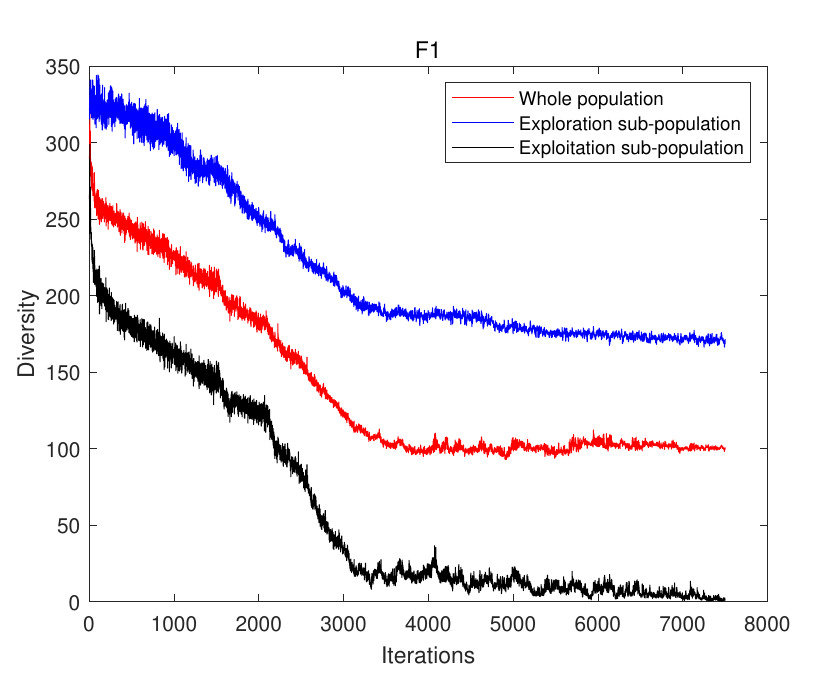}
	\end{minipage}
	}%
	\subfigure[F4 function]{
	\begin{minipage}[t]{0.45\linewidth}
	
	\includegraphics[scale=0.4]{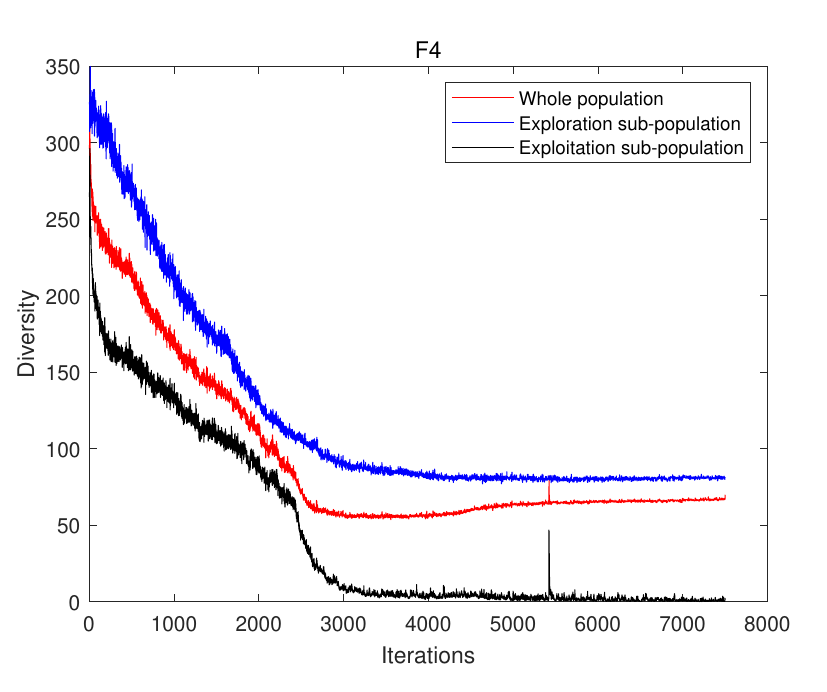}
	\end{minipage}%
	}%
	
	\subfigure[F17 function]{
	\begin{minipage}[t]{0.45\linewidth}
	
	\includegraphics[scale=0.4]{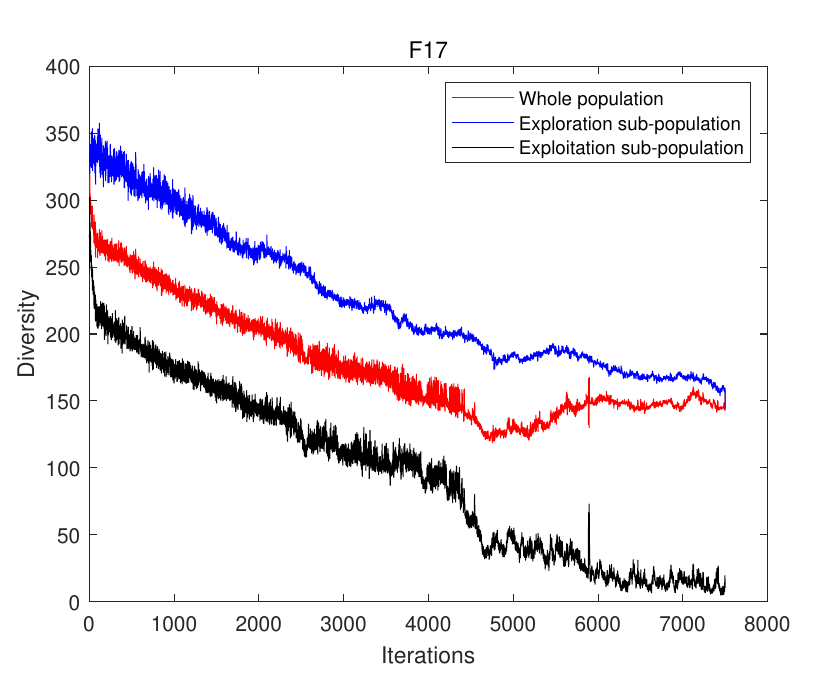}
	\end{minipage}
	}%
	\subfigure[F23 function]{
	\begin{minipage}[t]{0.45\linewidth}
	
	\includegraphics[scale=0.4]{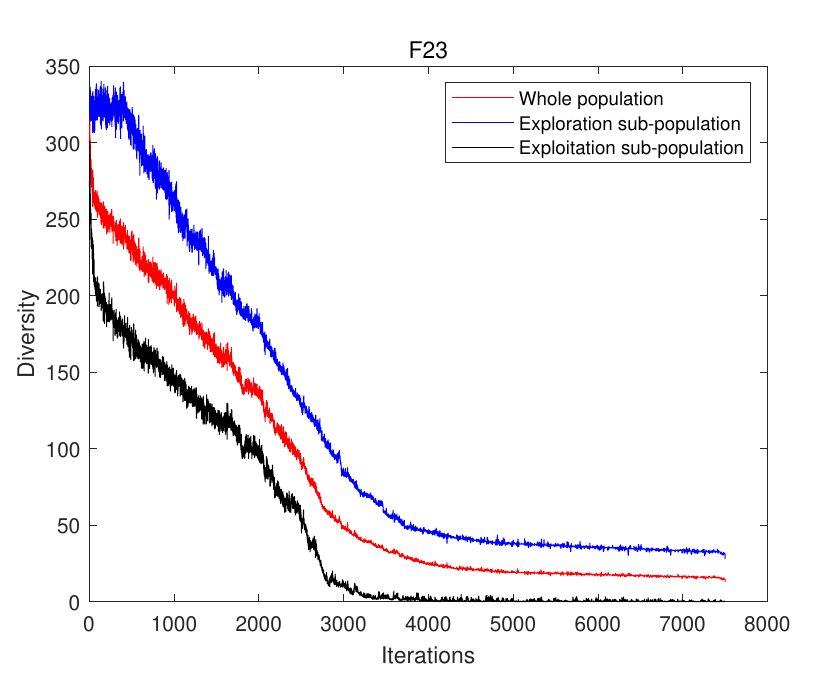}
	
	\end{minipage}
	}%

	\caption{Diversity comparisons between the exploration sub-population, the exploitation sub-population, and the entire population.}
	
	\label{fig7}
\end{figure}
To show the diversity of SpadePSO quantitatively, we select the CEC2014 benchmark suite, which comprises 30 test functions divided into four groups, namely three unimodal functions (F1 $ \sim $ F3), thirteen simple multimodal functions (F4 $ \sim $ F16), six hybrid functions (F17 $ \sim $ F22), and eight composite functions combing multiple test problems into a complex landscape (F23 $ \sim $ F30). The full CEC2014 benchmark suite is designed with a high difficulty level, because it involves composing test problems by extracting features dimension-wise from several problems, graded level of linkages, and rotated trap problems. In addition, the search range of the CEC2014 benchmark suite is $[-100,100]^D$. In Figure~\ref{fig7}, we present the diversity curves of F1, F4, F17, and F23 (the first function from each group), where the dimension of the search space, population size, and maximum number of fitness function evaluations are set to 30, 40, and 7500, respectively. As presented in Table~\ref{table2}, we list the minimum difference (Min), maximum difference (Max), averaged difference (Mean), and standard deviation (Std) between the exploration and the exploitation sub-populations across the entire optimization process. The results show that the exploration sub-population maintains the highest diversity consistently, which may allow SpadePSO to escape from the local trapping regions. Furthermore, the exploitation sub-population maintains the smallest diversity, which can fine-search in the currently optimal solution regions. As shown, the heterogeneous population design achieves our original intention.

\begin{table*}[!t]
\centering
\caption{Difference between the exploration and the exploitation sub-populations}
				\label{table2}
				\scalebox{0.8}{
		\begin{tabular}{|c|c|c|c|c|}
			\hline
			\diagbox[width=6.45em]{Stat.}{Function} &F1 &F4 &F17 &F23 \\ \hline
			Mean  & 161.05 & 55.92  & 143.17  &  56.29   \\ \hline
			Max   & 197.30 & 174.79  & 195.70  &  142.68    \\ \hline
			Min   & 27.00 & 11.96 & 19.39  & 6.88   \\ \hline
			Std   & 32.17 & 31.14  &26.30  & 34.63  \\ \hline
		\end{tabular}}
\end{table*}
\section{Performance Evaluations of SpadePSO with Comparisons}
\label{sec4}
In this section, we first introduce the experimental setups, and then present the determination of parameter values in SpadePSO. Subsequently, we evaluate the influence of various topologies on SpadePSO. Finally, we present the experimental results on solving the full CEC2014 benchmark suite,  CEC2013 large-scale benchmark suite, CEC2018 dynamic multi-objective optimization problem benchmark suite and two real-world optimization problems. 
 
\subsection{Experimental setups}
\label{sec4.0}

For all the experiments reported in this paper, we set the population size consistently to 40, the maximum value of velocity $v_{max}$ to 10\% of the search range, the number of runs per function to 30, and the maximum number of fitness function evaluations per run to 10000$\cdot D$ following the settings used in \citep{suganthan_problem_2005}. All experiments run on the same computer with an Intel Core i7 @ 2.90 GHz CPU and 16G memory. The parameter settings of all benchmarking algorithms are listed in Table~\ref{table3}.

\begin{table}[!t]
	\centering
	\caption{Parameters settings of the benchmarking algorithms}
	\label{table3}
	\scalebox{0.65}{
		\begin{tabular}{|l|l|l|}
			\hline
			& Algorithm   & Parameters settings                                                                        \\ \hline
			1. & PSO (1995) \citep{eberhart_new_1995}       & w : 0.9$\sim$0.4, $c_{1}$ = $c_{2}$ = 2                                                                  \\ \hline
			2. & CLPSO (2006) \citep{liang_comprehensive_2006}    & w : 0.9$\sim$0.4, $c_{1}$ = $c_{2}$ = 1, $c$ = 0.5                                                                    \\ \hline
			3.&OLPSO (2011) \citep{zhan_orthogonal_2011} 	 & w : 0.9$\sim$0.4, $c_{1}$ = $c_{2}$ = 1, $c$ = 0.5   \\ \hline
			4. & L-SHADE (2014) \citep{tanabe_improving_2014}   & $r^{N^{init}}$ = 18, $r^{arc}$ = 2.6, p = 0.11, H = 6        \\ \hline
			5. & HCLPSO (2015) \citep{lynn_heterogeneous_2015}    & \tabincell{c}{w : 0.99$\sim$0.2, $c_{1}$ : 2.5$\sim$0.5, $c_{2}$ : 0.5$\sim$2.5, \\c : 3$\sim$1.5}                                \\ \hline
			6. & GL-PSO (2016) \citep{gong_genetic_2016}    & \tabincell{c}{w = 0.7298, $c_{1}$ = $c_{2}$ = 1.49618, $C_{R}$ = 0.5, \\I = 4, $\delta$ = 0.2, $F_{i}\in{{[}-1,-0.4{]}\cup{[}0.4,0.1{]}}$ }           \\ \hline
			7. &TSLPSO (2019) \citep{xu_particle_2019}	 & w : 0.9$\sim$0.4, $c_{1}$ = $c_{2}$ = 1.5, $c_{3}$ : 0.5$\sim$2.5  \\ \hline
			8. &SPA-CatlePSO (2019) \citep{cui_surprisingly_2019} & \tabincell{c}{ w = 0.99$\sim$0.2, $c_{1}$ = 2.5$\sim$0.5, $c_{2}$ = 0.5$\sim$2.5 \\c : 3$\sim$1.5, $k$ = 8, $v_k$ = 8, $n_{exp}$ = 3}   \\ \hline
			9. &XPSO (2020) \citep{xia_expanded_2020} & $\eta$ = 0.5, $Stag_{max}$ = 5, $p$=0.2   \\ \hline
			10. &CC-RDG3 (2019) \citep{Sun_Decomposition_2019}  & ${\epsilon}_n$ = 50, ${\epsilon}_s$ = 100,\\ \hline
			11.& DMO (2022) \citep{agushaka_dwarf_2022} & babysitters=3, alpha group = scouts = 37, alpha female = 2 \\ \hline
			12. & SpadePSO (ours)& \tabincell{c}{ w = 0.99$\sim$0.2, $c_{1}$ = 2.5$\sim$0.5, $c_{2}$ = 0.5$\sim$2.5 \\c : 3$\sim$1.5, $k_0$ = 2, $v_k$ = 6, $n_{exp}$ = 5}        \\ \hline
		\end{tabular}}
	\end{table}

\subsection{Determination of parameter values in SpadePSO}
\label{sec4.1}

Other than the common PSO parameters, the performance of SpadePSO is also determined by three dedicated parameters, namely (i)~the out-degree $k_0$ (see Eq. \eqref{e18}), (ii)~the increasing velocity of out-degree $v_k$ (see Eq. \eqref{e18}), and (iii)~the number of expert particles $n_{exp}$ (see Eq. \eqref{e19}). If the values of $k_0$, $v_k$, and $n_{exp}$ are too large, the topology would be akin to a fully connected structure, which may result in the search getting trapped in the local minima regions. Conversely, if the values are too small, the topology would be akin to a ring structure, which does not easily converge \cite{kennedy_population_2002}. The common PSO parameters, including the population size of two sub-populations, inertia weight $w$, and acceleration coefficients $c$, $c_1$, and $c_2$, are adopted from those used in HCLPSO \citep{lynn_heterogeneous_2015} for fair comparisons. 

\begin{table}[!t]
\centering
	\caption{Determination of parameter values in SpadePSO (the lower the Ave. rank value, the better the performance of SpadePSO)}
	\label{table4}
	\scalebox{0.75}{
\begin{tabular}{c|c|c|c|c|c|c|c|c|c|c|c|c|c|c|c|c|c}
\hline
\multicolumn{18}{|l|} {Evaluation of $k_0 + v_{k}$ without $G\left(\bm{V}, \bm{E}_{exp}\right)$}                                                                                                                                                                                        \\ \hline
\multicolumn{3}{|l|}{$k_0$ + $v_{k}$ }            & \multicolumn{2}{l}{4}   & \multicolumn{1}{l|}{}          & \multicolumn{2}{l}{8}     & \multicolumn{1}{l|}{}         & \multicolumn{2}{l}{12}   & \multicolumn{1}{l|}{}            & \multicolumn{2}{l}{16}     & \multicolumn{1}{l|}{}         & \multicolumn{2}{l}{20}       & \multicolumn{1}{l|}{}        \\ \hline
\multicolumn{3}{|l|}{Ave. rank}             & \multicolumn{2}{l}{3.19}  & \multicolumn{1}{l|}{}           & \multicolumn{2}{l}{2.66}           & \multicolumn{1}{l|}{}   & \multicolumn{2}{l}{2.84}        & \multicolumn{1}{l|}{}       & \multicolumn{2}{l}{3.16}     & \multicolumn{1}{l|}{}       & \multicolumn{2}{l}{3.16}      & \multicolumn{1}{l|}{}       \\ \hline
\multicolumn{3}{|l|}{Final rank}             & \multicolumn{2}{l}{5}  & \multicolumn{1}{l|}{}          & \multicolumn{2}{l}{1}     & \multicolumn{1}{l|}{}       & \multicolumn{2}{l}{2}     & \multicolumn{1}{l|}{}        & \multicolumn{2}{l}{3}   & \multicolumn{1}{l|}{}         & \multicolumn{2}{l}{4}      & \multicolumn{1}{l|}{}       \\ \hline
\multicolumn{18}{|l|}{Evaluation of $k_0$ and $v_{k}$ without $G\left(\bm{V}, \bm{E}_{exp}\right)$}                                                                                                                                                                                        \\ \hline
\multicolumn{2}{|l|}{$k_0$} & \multicolumn{2}{l|}{1} & \multicolumn{2}{l|}{2} & \multicolumn{2}{l|}{3} & \multicolumn{2}{l|}{4} & \multicolumn{2}{l|}{5} & \multicolumn{2}{l|}{6} & \multicolumn{2}{l|}{7} & \multicolumn{2}{l|}{8} \\ \hline
\multicolumn{2}{|l|}{$v_{k}$} & \multicolumn{2}{l|}{7} & \multicolumn{2}{l|}{6} & \multicolumn{2}{l|}{5} & \multicolumn{2}{l|}{4} & \multicolumn{2}{l|}{3} & \multicolumn{2}{l|}{2} & \multicolumn{2}{l|}{1} & \multicolumn{2}{l|}{0} \\ \hline
\multicolumn{2}{|l|}{Ave. rank} & \multicolumn{2}{l|}{4.13} & \multicolumn{2}{l|}{3.88} & \multicolumn{2}{l|}{4.84} & \multicolumn{2}{l|}{4.49} & \multicolumn{2}{l|}{4.59} & \multicolumn{2}{l|}{5.19} & \multicolumn{2}{l|}{5.22} & \multicolumn{2}{l|}{5.22} \\ \hline
\multicolumn{2}{|l|}{Final rank} & \multicolumn{2}{l|}{3} & \multicolumn{2}{l|}{1} & \multicolumn{2}{l|}{6} & \multicolumn{2}{l|}{4} & \multicolumn{2}{l|}{5} & \multicolumn{2}{l|}{7} & \multicolumn{2}{l|}{8} & \multicolumn{2}{l|}{2} \\ \hline
\multicolumn{18}{|l|}{Evaluation of $n_{exp}$ with $k_0$ = 2, $v_{k}$ = 6}                                                                                                                                                                                        \\ \hline
\multicolumn{3}{|l|}{$n_{exp}$ }      & \multicolumn{2}{l}{2}     & \multicolumn{1}{l|}{}        & \multicolumn{2}{l}{3}      & \multicolumn{1}{l|}{}       & \multicolumn{2}{l}{4}  & \multicolumn{1}{l|}{}            & \multicolumn{2}{l}{5}    & \multicolumn{1}{l|}{}         & \multicolumn{2}{l}{6}        & \multicolumn{1}{l|}{}      \\\hline
\multicolumn{3}{|l|}{Ave. rank }             & \multicolumn{2}{l}{3.28}      & \multicolumn{1}{l|}{}       & \multicolumn{2}{l}{2.97}           & \multicolumn{1}{l|}{}  & \multicolumn{2}{l}{3.06}      & \multicolumn{1}{l|}{}        & \multicolumn{2}{l}{2.75}       & \multicolumn{1}{l|}{}      & \multicolumn{2}{l}{2.94}   & \multicolumn{1}{l|}{}           \\ \hline
\multicolumn{3}{|l|}{Final rank}             & \multicolumn{2}{l}{5}     & \multicolumn{1}{l|}{}        & \multicolumn{2}{l}{3}    & \multicolumn{1}{l|}{}         & \multicolumn{2}{l}{4}     & \multicolumn{1}{l|}{}         & \multicolumn{2}{l}{1}   & \multicolumn{1}{l|}{}          & \multicolumn{2}{l}{2}     & \multicolumn{1}{l|}{}  \\ \hline      
\end{tabular}}

\end{table}

The optimization problem is adopted from the first sixteen functions of the CEC2014 benchmark suite (see Section \ref{sec3.4} for more details) \citep{liang_problem_nodate}, whose $D$ is 30. We adopt the same parameter value  determination method as used in \citep{cui_surprisingly_2019}, i.e., we sequentially determine the out-degree value, i.e., $k_0 + v_{k}$, the values of $k_0$ and $v_{k}$, and the value of $n_{exp}$ by trial-and-error. As presented in Table~\ref{table4}, the results are ranked based on the mean of fitness values. According to the evaluation of $k_0 + v_{k}$, we set $k_0 + v_{k}$ to 8 to further determine the values of $k_0$ and $v_{k}$. According to the experimental results for various values of $k_0$ and $v_{k}$, we set $k_0$ and $v_{k}$ to 2 and 6, respectively. Finally, according to the evaluation of $n_{exp}$, we set $n_{exp}$ to 5. These parameter values are used for all subsequent experiments.

\subsection{Influence of various topologies on SpadePSO}
\label{sec4.2}
An appropriate topology may improve the performance of SpadePSO. To evaluate our proposed Euclidean distance-based topology, in Table~\ref{table5}, we compare the performance of SpadePSO with three different topologies on the full CEC2014 benchmark suite, with $D$ = 10, 30, 50, and 100. 

\begin{table}[!t]
	\centering
	\caption{Wilcoxon signed ranks test of three different topologies of SPA}
	\label{table5}
	\scalebox{0.75}{
		\begin{tabular}{|c|c|c|c|c|c|}
			\hline
			\tabincell{c}{First topology (SpadePSO)\\ VS.}          & \diagbox[width=6em]{Difference}{Dimension} & {10D}       & {30D}       & {50D}        & {100D}      \\ \hline
					{}                          & {+}    & {13}        & {15}  & {17}         & {19}        \\ \cline{2-6} 
					{}                          & {-}    & {15} 	   & {14}         & {12}         & {10}        \\ \cline{2-6} 
					{}							& {$\approx$}    & {2}  & {1}          & {1}    & {1}       \\ \cline{2-6}
	           \multirow{-4}{*}{{Second topology}}   & {$p$}    & {0.75}  & {0.56}          & {0.36}    & {0.21} \\ \hline
					{}                          & {+}    & {13}        & {15}         & {16}         & {17}        \\ \cline{2-6} 
					{}                          & {-}    & {15}        & {14}         & {13}         & {11}        \\ \cline{2-6} 
					{}                          & {$\approx$}    & {2}  & {1}   & {1}    & {2}                   \\ \cline{2-6} 
	           \multirow{-4}{*}{{Third topology}}  & {$p$}    & {0.82}  & {0.74}   & {0.84}    & {0.08}  \\ \hline
		\end{tabular}}
	\end{table}
	
The first topology is the one adopted by SpadePSO (see Section \ref{sec3.1}). In the initial iteration, the connections between particles are merely determined by the Euclidean distance. During each subsequent iteration, other than the Euclidean distance information, particles with high fitness values are also used to construct the contemporary topology. The second topology is the one used in SPA-CatlePSO \cite{cui_surprisingly_2019}, i.e.,  Particle $i$ is unidirectionally connected to particles numbered from $i+1$ to $i+k$ in the initial iteration. During each subsequent iteration, particles with high fitness values are used to construct the contemporary topology. In addition, if Particle $i$ selects $\bm{x}^{sbest}$ as the learning exemplar and its fitness value is improved in the current iteration, Particle $i$ is connected to the particle corresponding to $\bm{x}^{ {sbest}}$ in the subsequent iteration. In the third topology, we combine the first and second topologies such that the connections between particles are determined by the Euclidean distance in the initial iteration. During the iteration, the topology is constructed in the same way as the first topology. Furthermore, the particle corresponding to $\bm{x}^{ {sbest}}$ is connected in the same way as the second topology.

To better quantitatively assess the experimental results, we adopt the well-known, widely-used nonparametric statistics analysis Wilcoxon signed ranks test \citep{derrac_practical_2011}. In Table~\ref{table5}, the symbols +, -, and $\approx$ indicate that the first topology performs significantly better (+), significantly worse (-), or not significantly different ($\approx$) compared to another topology. An algorithm can be considered significantly better than another if $p \leq$ 0.1. As presented in Table~\ref{table5}, the first topology performs best on 30, 50, and 100 dimensions but worst on 10 dimensions. Among the three topologies being evaluated, SpadePSO performs best on higher dimensionality.

\subsection{Experimental results on CEC2014 benchmark suite}
\label{sec4.3}
		
To assess the performance of SpadePSO, we conduct experiments on the full CEC2014 benchmark suite, with $D$ = 10, 30, 50, and 100. Table~\ref{table6} presents the Wilcoxon signed ranks test of SpadePSO and all benchmarking algorithms, where the results are compared based on the mean of fitness values. The Best, Mean, and Std of fitness values of SpadePSO are presented in Appendix A. 
			
\begin{table}[!t]
			\centering
			\caption{Wilcoxon signed ranks test of the CEC2014 benchmark suite}
			\label{table6}
			\scalebox{0.6}{
				\begin{tabular}{|c|c|c|c|c|c|c|c|c|c|c|c|}
					\hline
					\tabincell{c}{SpadePSO (ours) \\ VS.}          & \diagbox[width=6em]{Difference}{Dimension} & {10D}       & {30D}       & {50D}        & {100D}  &   \tabincell{c}{SpadePSO (ours) \\ VS.}          & \diagbox[width=6em]{Difference}{Dimension} & {10D}       & {30D}       & {50D}        & {100D} \\ \hline
					{}                          & {+}    & {29}  & {23 }  & {28 }   & {28 }  &{}                          & {+}    & {29 }  & {24}  & {26}   & {24 }\\ \cline{2-6} \cline{8-12} 
					{}                          & {-}    & {1}          & {7}          & {2}          & {2}     &{}                          & {-}    & {1}     & {6}  & {4}    & {6}   \\ \cline{2-6} \cline{8-12} 
					{}                 & {$\approx$}    & {0}          & {0}          & {0}          & {0} &{}  & {$\approx$}                    & {0}    & {0}   & {0}    & {0}\\ \cline{2-6} \cline{8-12} 
		    \multirow{-4}{*}{{PSO (1995)}}     & {$p$-value}    & {0.00}          & {0.01}          & {0.00}          & {0.00}    &\multirow{-4}{*}{{GL-PSO (2016)}}  & {$p$-value}    & {0.00}          & {0.00}          & {0.00}          & {0.00}        \\ \hline
		    		{}                          & {+}    & {19 }  & {15}         & {15}         & {17}     &{}                          & {+}    & {17}         & {19}         & {15}         & {18 }   \\ \cline{2-6} \cline{8-12} 
					{}                          & {-}    & {11}         & {15}         & {14}         & {13}    &{}                          & {-}    & {13} 	    & {11}         & {14}         & {12}      \\ \cline{2-6} \cline{8-12} 
					{}   & {$\approx$}    & {0}          & {0 }   & {1}  & {0} &{}   & {$\approx$}    & {0}   & {0 }   & {1}   & {0 } \\ \cline{2-6} \cline{8-12} 
			\multirow{-4}{*}{{CLPSO (2006)}}   & {$p$-value}    & {0.08}          & {0.88}   & {0.58}  & {0.88}  &\multirow{-4}{*}{{TSLPSO (2019)}}  & {$p$-value}    & {0.73}   & {0.19}   & {0.19}   & {0.09 }\\ \hline
					{}                          & {+}    & {25 }  & {19}  & {22 }  & {20} &{}                          & {+}    & {13}         & {15}         & {17}         & {19 }\\ \cline{2-6} \cline{8-12} 
					{}                          & {-}    & {5}          & {11}         & {7}          & {10}     &{}                          & {-}    & {15} 	    & {14}         & {12}         & {10}    \\ \cline{2-6} \cline{8-12} 
					{}   & {$\approx$}    & {0}          & {0}          & {1}          & {0}  &{}   & {$\approx$}    & {2}   & {1 }   & {1}   & {1 }\\ \cline{2-6} \cline{8-12} 
		    \multirow{-4}{*}{{OLPSO (2011)}}   & {$p$-value}    & {0.00}          & { 0.01}          & {0.00} & {0.02}     &\multirow{-4}{*}{{SPA-CatlePSO (2019)}}  & {$p$-value}    & {0.75}   & {0.56}   & {0.36}   & {0.21}    \\ \hline
					{}                          & {+}    & {4}          & {1}          & {2}          & {4}       &{}                          & {+}    & {25}   & {26}   & {27}   & {25}  \\ \cline{2-6} \cline{8-12} 
					{}                          & {-}    & {25}  & {28 }  & {24 }   & {26 }  &{}                          & {-}    & {4} 	        & {4}          & {3}          & {5} \\ \cline{2-6} \cline{8-12} 
					{}    & {$\approx$}    & {1}          & {1}          & {4}       & {0} &{}  & {$\approx$}    & {1}          & {0}          & {0 }         & {0 }\\ \cline{2-6}  \cline{8-12} 
			\multirow{-4}{*}{{L-SHADE (2014)}} & {$p$-value}    & {0.00}          & {0.00}          & {0.00}          & {0.00}    &\multirow{-4}{*}{{XPSO(2020)}}    &  {$p$-value}  & {0.00}          & {0.00}          & {0.00 }         & {0.00}     \\ \hline
					{}                          & {+}    & {25}  & {20}         & {21 }  & {19}  &{}     &{+}&{29}&{22}&{22}&{27} \\ \cline{2-6} \cline{8-12}
					{}                          & {-}    & {5}          & {10}         & {8}          & {10}  &{}  &{-}&{1}&{8}&{8}&{3}    \\ \cline{2-6} \cline{8-12}
					{}     & {$\approx$}    & {0 }         & {0}   & {1}          & {1} &{} &{$\approx$}&{0}&{0}&{0}&{0} \\ \cline{2-6} \cline{8-12}
			\multirow{-4}{*}{{HCLPSO (2015)}}  & {$p$-value}    & {0.01}         & {0.32}   & {0.05}          & {0.27}& \multirow{-4}{*}{{DMO (2022)}} &{$p$-value}&{0.00}&{0.00}&{0.01}&{0.00} \\ \hline
					 
				\end{tabular}}
			\end{table}	
			
In this subsection, the benchmarking algorithms selected for performance comparison include PSO \citep{eberhart_new_1995}, TSLPSO \citep{xu_particle_2019}, SPA-CatlePSO \citep{cui_surprisingly_2019}, HCLPSO \citep{lynn_heterogeneous_2015}, OLPSO \citep{zhan_orthogonal_2011}, GL-PSO \citep{gong_genetic_2016}, XPSO \citep{xia_expanded_2020}, DMO \citep{agushaka_dwarf_2022},  CLPSO \citep{liang_comprehensive_2006},  and  L-SHADE \citep{tanabe_improving_2014}. HCLPSO, TSLPSO, and XPSO are algorithms that improve the neighborhood topology of the population. HCLPSO and TSLPSO are heterogeneous PSO algorithms, with two separate sub-populations responsible for exploration and exploitation, respectively. In XPSO, particles learn locally and globally from populations' historical best position and dynamically update the topology. CLPSO uses CLS, which leads to a wider exploration. OLPSO searches for the best combination of $\bm{x}^{gbest}$ and $\bm{x}^{pbest}$ to construct the learning exemplars. L-SHADE was the best algorithm of the CEC2014 benchmark suite \cite{xu_particle_2019}. In L-SHADE, Linear Population Size Reduction is adopted, which reduces the population size linearly during the iteration to better balance between exploration and exploitation. GL-PSO combines GA and PSO to construct the learning exemplars. DMO is a novel metaheuristic algorithm consisting of three social groups, which mimics the foraging behavior of the dwarf mongoose.

	As presented in Table~\ref{table6}, SpadePSO performs significantly better than PSO, OLPSO, GL-PSO, XPSO, and DMO for 10, 30, 50, and 100 dimensions. Although the difference between SpadePSO and HCLPSO is not statistically significant, SpadePSO shows better results than HCLPSO on most of the test functions for 30 and 100 dimensions. SpadePSO has an exploitation sub-population that leads to finer exploitation. This may explain why SpadePSO performs better than CLPSO for 10 dimensions with a $p$-value of 0.08. With the increase of dimensions, SpadePSO performs better than TSLPSO and SPA-CatlePSO, which is further confirmed in the following subsection on the CEC2013 large-scale benchmark suite. As the Linear Population Size Reduction is adopted in L-SHADE, the initial population size of L-SHADE is much larger than that of PSO variants, i.e., 180, 540, 900, and 1800 on the 10, 30, 50, and 100 dimensions, respectively. This may explain why L-SHADE performs significantly better than all PSO variants.

\begin{figure}[!t]
	\centering
	\subfigure[F1 function]{
	\begin{minipage}[t]{0.45\linewidth}
	
	\includegraphics[scale=0.2]{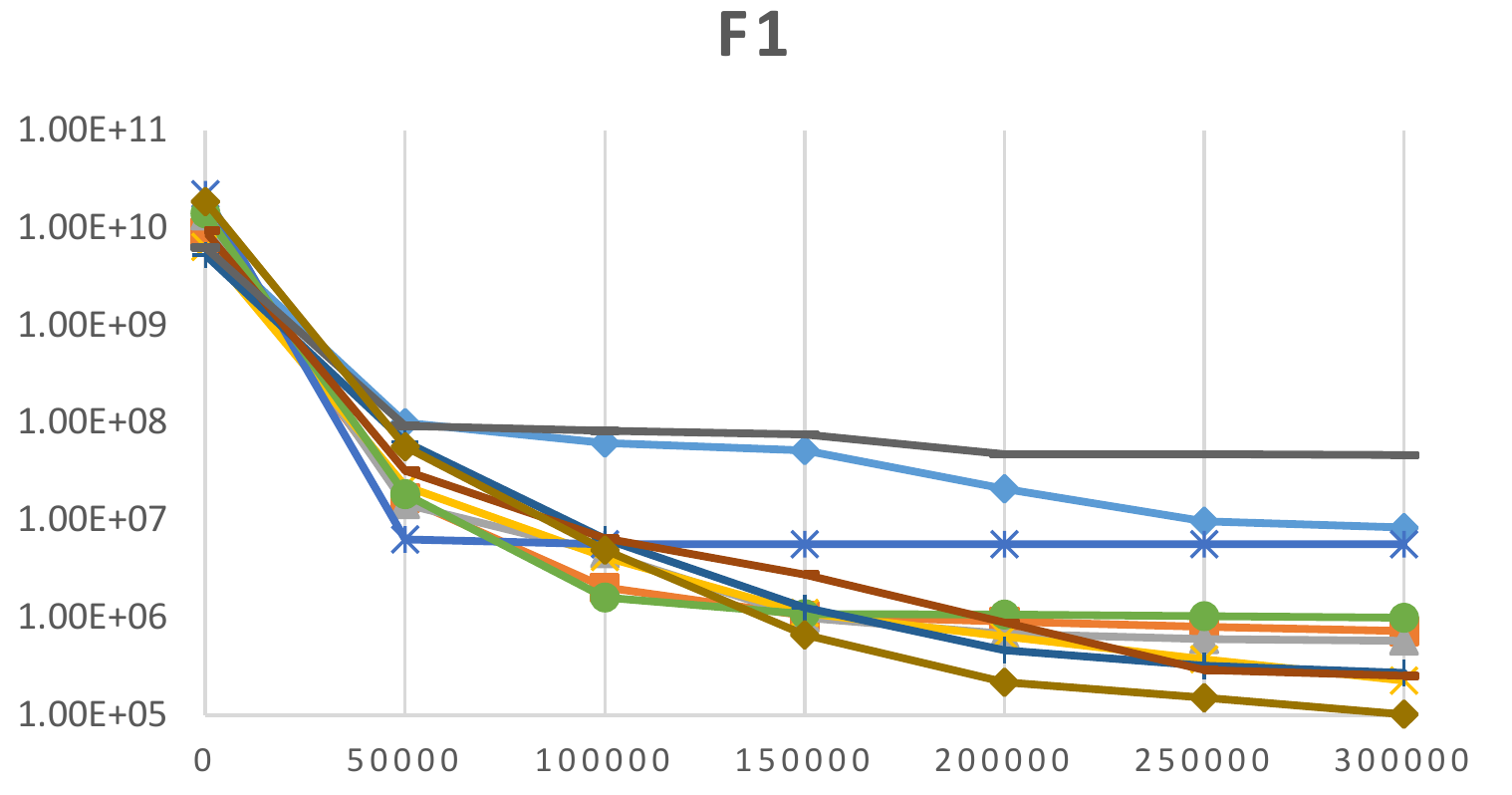}
	\end{minipage}
	}%
	\subfigure[F4 function]{
	\begin{minipage}[t]{0.45\linewidth}
	
	\includegraphics[scale=0.2]{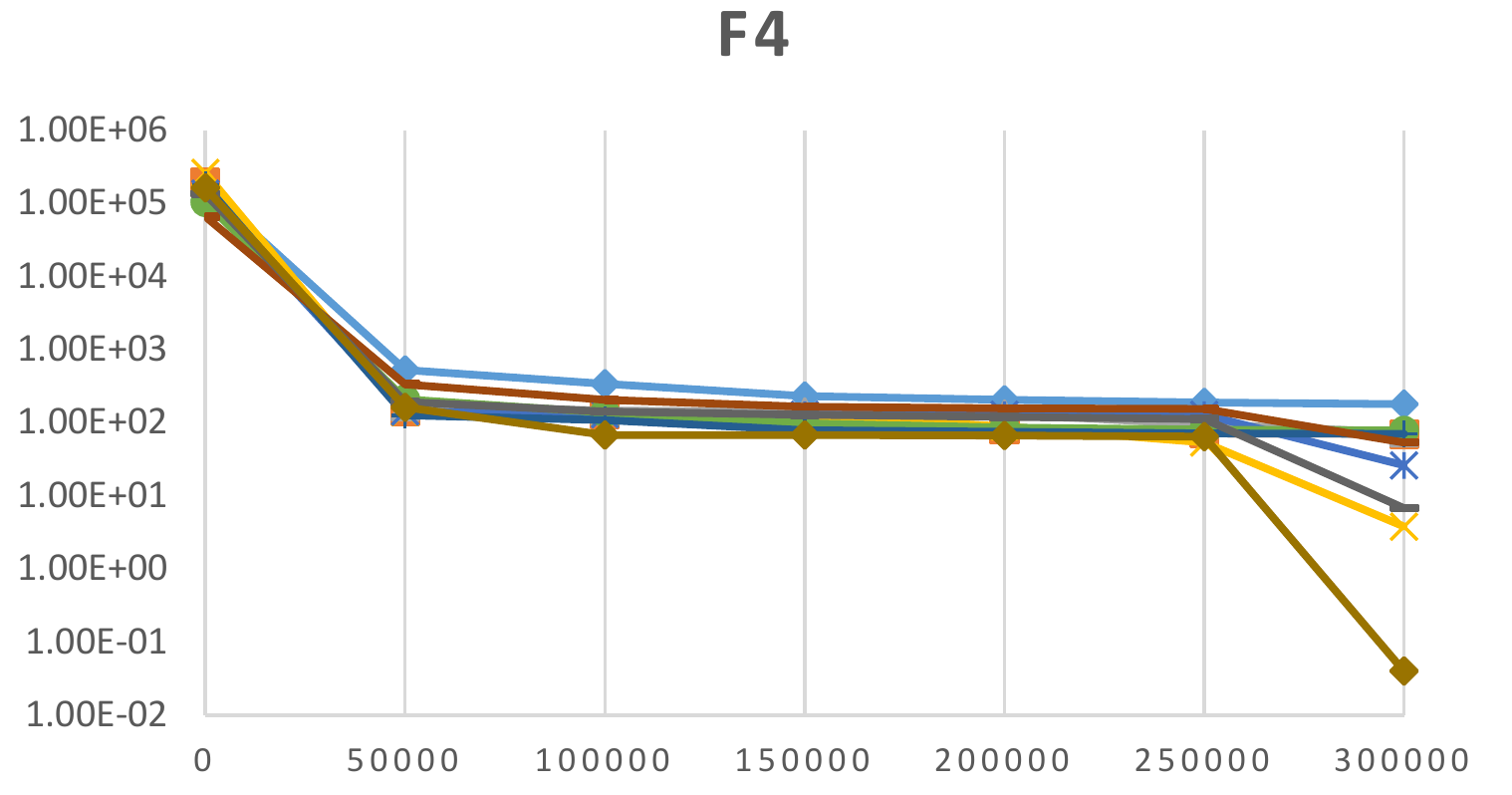}
	\end{minipage}%
	}%
	
	\subfigure[F17 function]{
	\begin{minipage}[t]{0.45\linewidth}
	
	\includegraphics[scale=0.2]{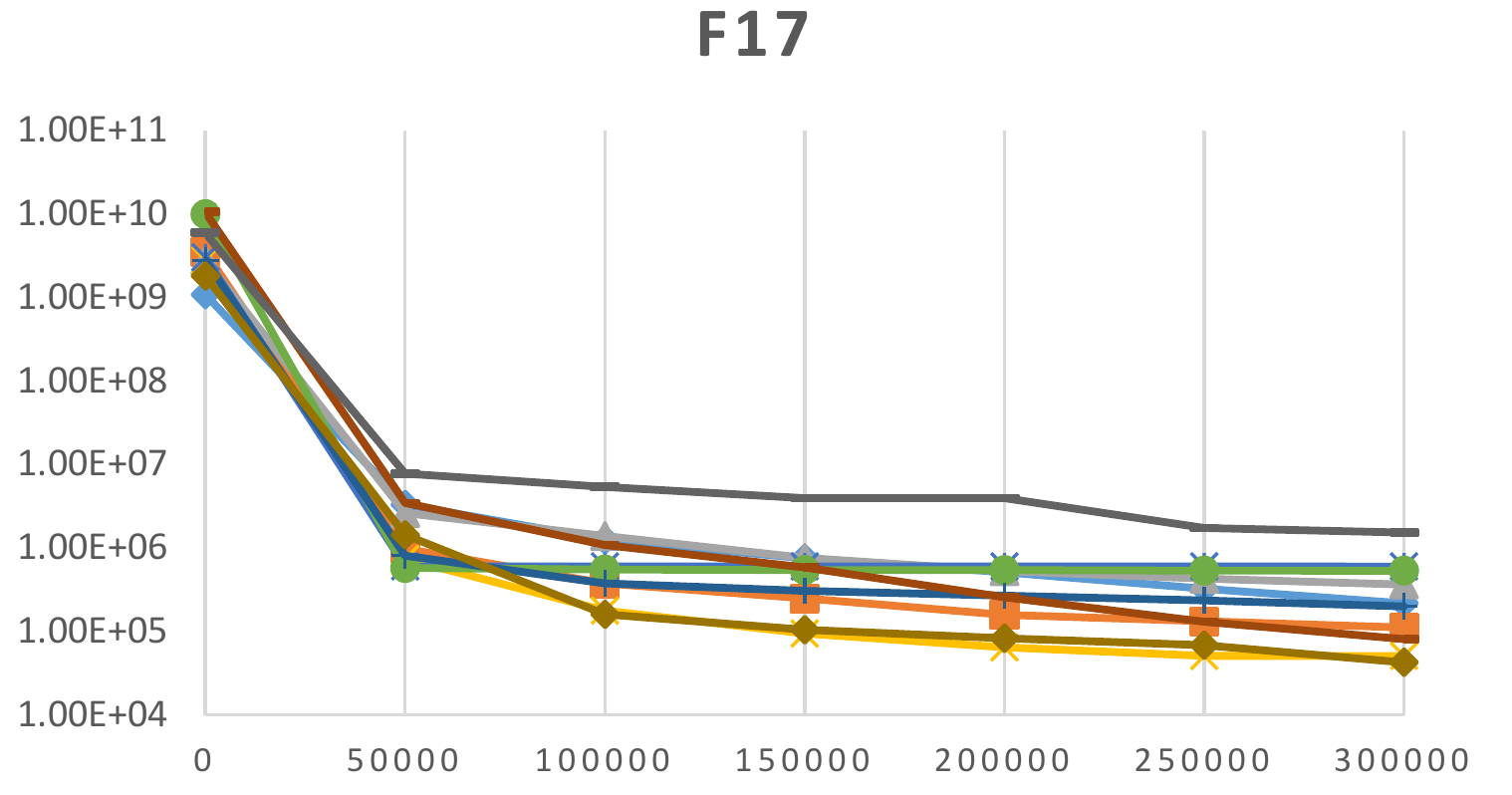}
	\end{minipage}
	}%
	\subfigure[F23 function]{
	\begin{minipage}[t]{0.45\linewidth}
	
	\includegraphics[scale=0.2]{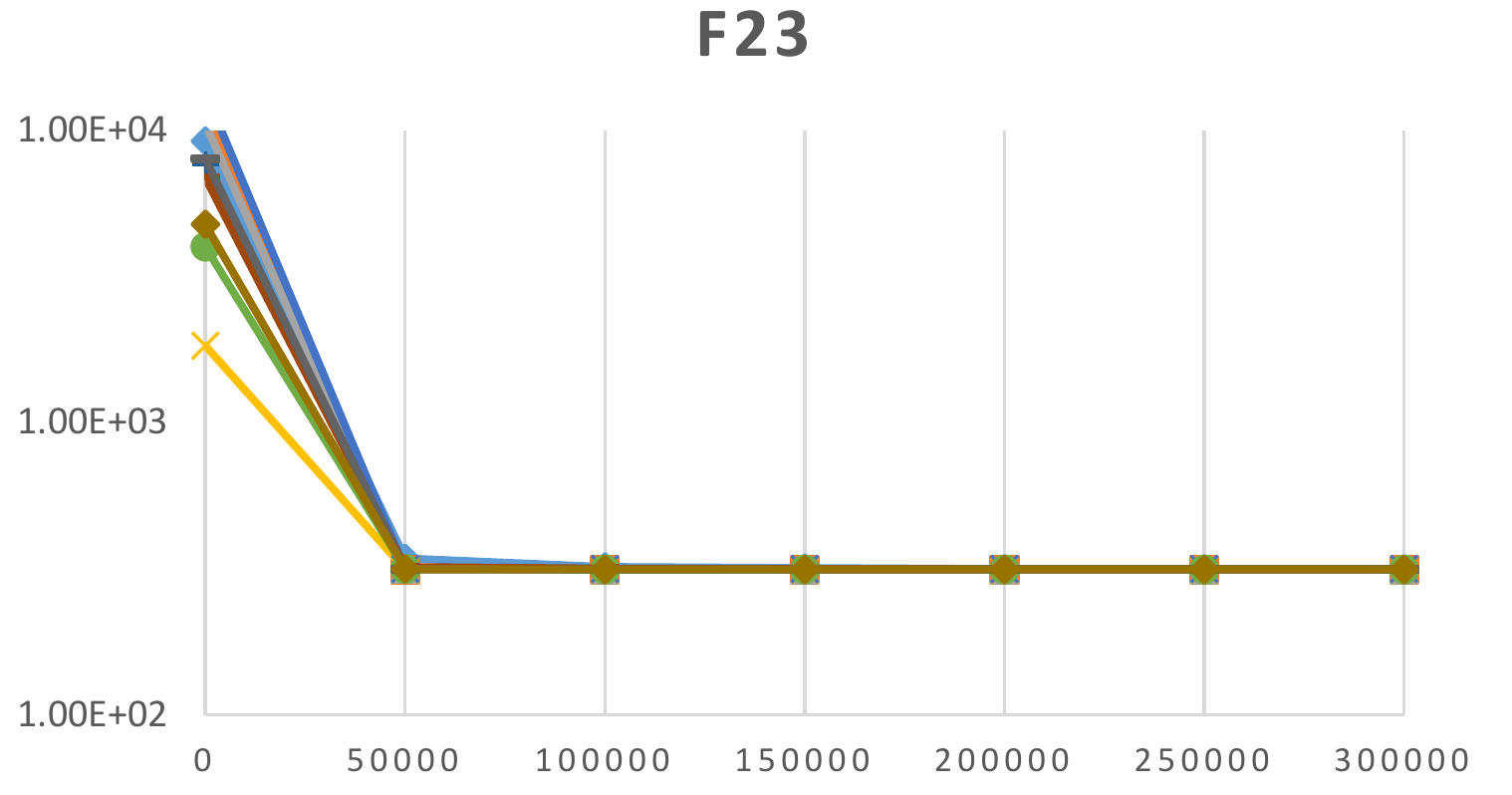}
	
	\end{minipage}}%

	\subfigure[legend]{
	\centering
	\begin{minipage}[t]{1\linewidth}
	
	\includegraphics[scale=0.5]{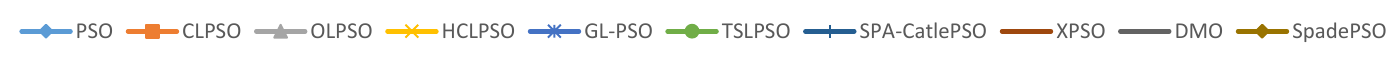}
	
	\end{minipage}
	}%
	
	\caption{Convergence curves of different algorithms.}
	\label{fig8}
\end{figure}

In Figure~\ref{fig8}, we present the convergence curves of F1, F4, F17, and F23 (the first function from each group) for 30 dimensions, to investigate why SpadePSO performs better than the other PSO variants. The horizontal axis represents the number of fitness evaluations, and the vertical axis represents the fitness value. For the F1 function, it can be clearly seen in Figure~\ref{fig8} that OLPSO, GLPSO, and TSLPSO fall into the local minima. In addition, it seems difficult for PSO and DMO to converge quickly, after 1E+5 fitness function evaluations. For the F4 function, all algorithms fall into the local minima for a long time. After 2.5E+5 fitness function evaluations, only SpadePSO, HCLPSO, and DMO escape from the local minima. In the end, SpadePSO finds a much better solution than the other algorithms do. For the F17 function, many algorithms fall into the local minima as with the F1 function. For the F23 function, all algorithms find the global optimal solution. As shown in Figure~\ref{fig8}, the convergence curves well demonstrate that SpadePSO is better at escaping local minima and producing better solutions. 

%In addition, we also admit that SpadePSO does not converge the fastest among all algorithms.

%Most of the multimodal functions contain a number of local optima, which may lead to premature convergence of PSO algorithms. It is difficult for the traditional PSO to locate the global optimum of the function F6, because this problem has many deep local optima far away from the global optima. Once a particle of the classical PSO is trapped into a deep local optimum, it could hardly escape.  As shown in Table 2, TSLPSO, CLPSO, and GL-PSO have better ability to resist the trapping of local optima and hence, reach the high accuracy of

\subsection{Experimental results on CEC2013 large-scale benchmark suite}
\label{sec4.4}
To assess the performance of SpadePSO on higher-dimensionality problems, we conduct experiments on the CEC2013 large-scale benchmark suite, whose $D$ is 905 (F13, F14) or 1000 (F1 $ \sim $  F12, F15). The benchmark suite consists of fifteen test functions divided into four groups, three fully-separable functions (F1 $ \sim $ F3), eight partially-separable functions (F4 $ \sim $ F11), three overlapping functions (F12 $ \sim $ F14), and a non-separable function (F15). In the field of continuous optimization, interaction between variables is commonly referred to as non-separability. Non-separability is one of the reasons why large-scale optimization problems are challenging for many algorithms. In addition, the overlapping implication is that the subcomponent functions share certain decision variables. Similarly, the overlapping problem is also partially-separable. 

In this subsection, other than PSO and its variants, we select CC-RDG3 \citep{Sun_Decomposition_2019} as the benchmarking algorithm for performance comparison. CC-RDG3 was designed for large-scale problems with overlapping subcomponents and is also the best algorithm of the CEC2013 large-scale benchmark suite. To solve overlapping problems, CC-RDG3 modifies the Recursive Differential Grouping (RDG) method and subdivides a large-scale problem into a set of smaller subproblems. We compare the Mean and Std of fitness values of all algorithms for comparison in Table~\ref{table7}. We also present the Wilcoxon signed ranks test of SpadePSO and all benchmarking algorithms in Table~\ref{table8}.

\begin{table}[!t]
			\centering
			\caption{Comparison of SpadePSO with other algorithms on the CEC2013 large-scale benchmark suite}
			\label{table7}
			\scalebox{0.45}{
				\begin{tabular}{|c|c|c|c|c|c|c|c|c|c|c|c|c|}
					\hline
					\tabincell{c}{Function}          & \diagbox[width=6em]{Stat.}{Algorithm} & \tabincell{c}{PSO\\(1995)}     & \tabincell{c}{CLPSO\\(2006)}  & \tabincell{c}{OLPSO\\(2011)} & \tabincell{c}{HCLPSO\\(2015)}   &\tabincell{c}{GL-PSO\\(2016)} & \tabincell{c}{TSLPSO\\(2019)} & \tabincell{c}{SPA-CatlePSO\\(2019)}& \tabincell{c}{XPSO\\(2020)} & \tabincell{c}{CC-RDG3\\(2019)} & \tabincell{c}{DMO\\(2022)}   & \tabincell{c}{SpadePSO\\(ours)} \\ \hline
					{} &{Mean} &{5.79E+08} &{4.17E+10} &{0.00E+00} &{0.00E+00} &{3.17E+08} &{2.41E-18} &{9.92E-09} &{1.22E+09} &{1.23E-18} &{3.39E+11} &{9.65E-09} \\ \cline{2-13} 
\multirow{-2}{*}{{F1}} & {Std} &{1.04E+08} &{3.88E+09} &{0.00E+00} &{0.00E+00} &{3.88E+07} &{3.41E-18} &{2.99E-11} &{1.02E+08} &{7.26E-20} &{6.45E+09} &{1.80E-10} \\ \hline

					{} &{Mean} &{3.62E+03} &{2.75E+04} &{3.89E+02} &{2.39E+00} &{2.74E+03} &{9.95E-01} &{3.48E+00} &{6.28E+03} &{2.35E+03} &{1.12E+05} &{1.99E+00} \\ \cline{2-13} 
\multirow{-2}{*}{{F2}} & {Std} &{4.43E+02} &{4.58E+02} &{4.20E+00} &{1.51E+00} &{1.03E+02} &{1.18E-12} &{2.11E+00} &{5.78E+02} &{ 9.78E+01} &{1.24E+03} &{1.09E+00} \\ \hline

					{} &{Mean} &{1.47E+01} &{1.97E+01} &{4.78E+00} &{1.19E-12} &{8.41E+00} &{5.22E-10} &{9.95E-09} &{1.06E+01} &{2.00E+01} &{2.15E+01} &{9.90E-09}  \\ \cline{2-13} 
\multirow{-2}{*}{{F3}} & {Std} &{5.65E+00} &{2.14E-02} &{2.36E-01} &{9.13E-14} &{1.33E-01} &{5.95E-10} &{3.27E-11} &{1.71E+00} &{1.15E-02}         &{ 3.00E-03} &{9.33E-11} \\ \hline

					{} &{Mean} &{7.09E+10} &{5.48E+11} &{7.78E+10} &{2.61E+09} &{2.62E+11} &{1.52E+11} &{1.98E+09} &{5.76E+10} &{3.15E-18} &{1.43E+13} &{2.59E+09}  \\ \cline{2-13} 
\multirow{-2}{*}{{F4}} & {Std} &{4.46E+10} &{1.27E+11} &{4.20E+10} &{1.37E+09} &{1.07E+11} &{1.13E+10} &{ 6.06E+08} &{ 2.69E+10 } &{2.30E-19} &{5.79E+11} &{5.55E+08} \\ \hline

					{} &{Mean} &{3.98E+06} &{4.61E+06} &{7.46E+06} &{1.88E+07} &{5.00E+06} &{7.61E+06} &{2.17E+07} &{2.28E+06} &{2.29E+06} &{6.27E+07} &{2.00E+07} \\ \cline{2-13} 
\multirow{-2}{*}{{F5}} & {Std} &{9.42E+05} &{4.91E+05} &{1.50E+06} &{3.30E+06} &{1.13E+06} &{1.05E+06} &{2.77E+06} &{4.67E+05} &{7.76E+04} &{3.57E+06} &{2.57E+06} \\ \hline

					{} &{Mean} &{2.11E+05} &{6.13E+05} &{1.65E+05} &{9.84E+05} &{2.09E+05} &{4.98E+05} &{9.84E+05} &{1.74E+05} &{9.96E+05} &{1.06E+06} &{3.83E+05} \\ \cline{2-13} 
\multirow{-2}{*}{{F6}} & {Std} &{9.91E+03} &{5.85E+04} &{5.98E+04} &{1.64E+03} &{1.66E+04} &{5.40E+05} &{3.40E+03} &{1.97E+04} &{2.64E-01} &{3.93E+04} &{8.33E+04} \\ \hline

					{} &{Mean} &{6.92E+08} &{1.58E+10} &{1.03E+09} &{4.40E+04} &{1.23E+09} &{3.50E+08} &{1.17E+05} &{5.11E+08} &{1.44E-21} &{1.06E+06} &{5.50E+04} \\ \cline{2-13} 
\multirow{-2}{*}{{F7}} & {Std} &{9.20E+08} &{5.04E+09} &{3.02E+08} &{1.12E+04} &{3.37E+08} &{2.31E+07} &{2.19E+04} &{4.55E+08} &{1.27E-22} &{7.36E+14} &{9.96E+03} \\ \hline

					{} &{Mean} &{5.01E+14} &{3.57E+15} &{3.56E+16} &{6.56E+13} &{7.07E+14} &{3.93E+15} &{7.10E+13} &{1.67E+14} &{6.03E+03} &{1.25E+18} &{6.79E+13} \\ \cline{2-13} 
\multirow{-2}{*}{{F8}} & {Std} &{2.28E+14} &{3.15E+15} &{2.65E+16} &{2.73E+13} &{5.49E+14} &{9.35E+14} &{9.53E+12} &{1.27E+15} &{1.89E+03} &{2.48E+17} &{1.70E+13} \\ \hline

					{} &{Mean} &{4.11E+08} &{4.75E+08} &{4.87E+08} &{1.31E+09} &{3.31E+08} &{5.01E+08} &{1.01E+09} &{2.67E+08} &{1.76E+08} &{4.33E+09} &{1.18E+09} \\ \cline{2-13} 
\multirow{-2}{*}{{F9}} & {Std} &{7.48E+07} &{6.84E+07} &{6.69E+07} &{1.27E+08} &{1.27E+08} &{2.78E+07} &{4.20E+08} &{6.79E+07} &{4.71E+06} 
&{5.98E+08} &{3.27E+08} \\ \hline

					{} &{Mean} &{1.04E+07} &{1.75E+07} &{2.71E+07} &{8.83E+07} &{1.13E+07} &{5.86E+07} &{8.87E+07} &{7.11E+06} &{9.05E+07} &{9.35E+07} &{4.29E+07} \\ \cline{2-13} 
\multirow{-2}{*}{{F10}} & {Std} &{1.25E+06} &{7.23E+06} &{5.31E+06} &{1.39E+06} &{2.84E+06} &{1.74E+07} &{6.57E+05} &{8.29E+06} &{2.83E+01} &{9.68E+05} &{3.72E+07} \\ \hline

					{} &{Mean} &{1.57E+09} &{4.78E+12} &{1.57E+11} &{8.35E+07} &{3.26E+11} &{8.62E+10} &{7.65E+07} &{7.00E+09} &{1.57E-19} &{1.19E+17} &{4.38E+07} \\ \cline{2-13} 
\multirow{-2}{*}{{F11}} & {Std} &{6.64E+08} &{1.63E+12} &{1.01E+11} &{5.98E+07} &{1.59E+11} &{1.11E+10} &{2.86E+07} &{1.22E+09} &{2.86E-20} &{5.26E+16} &{2.19E+07} \\ \hline

					{} &{Mean} &{7.26E+08} &{7.30E+11} &{3.74E+01} &{2.45E+01} &{7.64E+08} &{8.76E+02} &{1.98E+01} &{6.73E+08} &{2.62E+02} &{8.05E+12} &{2.11E+01} \\ \cline{2-13} 
\multirow{-2}{*}{{F12}} & {Std} &{2.19E+07} &{9.71E+10} &{4.90E+01} &{3.33E+00} &{1.54E+08} &{4.26E+01} &{3.53E+00} &{1.40E+07} &{1.02E+02} &{3.66E+10} &{2.50E+00} \\ \hline

					{} &{Mean} &{1.02E+10} &{3.67E+11} &{1.49E+10} &{5.88E+07} &{1.30E+10} &{1.90E+10} &{4.15E+07} &{1.70E+10}  &{1.90E+04}       &{1.68E+17} &{3.12E+07} \\ \cline{2-13} 
\multirow{-2}{*}{{F13}} & {Std} &{3.12E+09} &{3.40E+11} &{1.61E+09} &{3.22E+07} &{1.40E+09} &{1.80E+10} &{6.04E+06} &{6.13E+09} &{4.47E+02} &{2.67E+16} &{4.58E+06} \\ \hline

					{} &{Mean} &{4.63E+10} &{5.21E+12} &{2.42E+11} &{2.50E+07} &{1.91E+11} &{2.91E+11} &{2.29E+07} &{1.03E+11}  &{2.37E+09}       &{1.96E+17} &{2.70E+07} \\ \cline{2-13} 
\multirow{-2}{*}{{F14}} & {Std} &{6.40E+09} &{1.69E+12} &{5.80E+10} &{2.91E+06} &{6.44E+10} &{5.71E+10} &{3.68E+06} &{4.78E+10} &{3.29E+09} &{5.90E+16} &{8.04E+05} \\ \hline

					{} &{Mean} &{1.33E+07} &{3.00E+08} &{ 5.03E+07} &{8.62E+07} &{2.71E+07} &{1.52E+07} &{5.04E+07} &{1.93E+07}  &{7.74E+04}       &{3.14E+17} &{2.18E+06} \\ \cline{2-13} 
\multirow{-2}{*}{{F15}} & {Std} &{1.24E+06} &{4.21E+07} &{1.23E+07} &{1.79E+07} &{5.61E+06} &{8.72E+06} &{6.95E+07} &{1.77E+06} &{2.29E+04} &{1.68E+16} &{4.23E+05
} \\ \hline
				\end{tabular}}
			\end{table}

As presented in Table~\ref{table7}, on fully-separable functions (F1 $ \sim $ F3), HCLPSO, TSLPSO, Spa-CatlePSO, and SpadePSO perform better than the other PSO variants. It is worth noting that they all adopt a heterogeneous population topology. This may be the direction to solving large-scale fully-separable functions. For partially-separable (F4 $ \sim $ F11) and overlapping problems (F12 $ \sim $ F14), TSLPSO performs worse than HCLPSO, Spa-CatlePSO and SpadePSO on many functions. On the non-separable function (F15), SpadePSO performs significantly better than all PSO variants, including HCLPSO, TSLPSO, and Spa-CatlePSO. It is thus shown that SpadePSO is better suited to solve for non-separable functions than other PSO variants. As presented in Table~\ref{table8}, SpadePSO performs significantly better than PSO, CLPSO, OLPSO, GL-PSO, TSLPSO, XPSO, and DMO. CC-RDG3 is specially designed to solve large-scale problems by breaking the linkage among variables shared by multiple subcomponents. It is extremely encouraging to see that the performance of SpadePSO is not significantly different from that of CC-RDG3 ($p$=0.23).

\begin{table}[!t]
	\small
	\caption{Wilcoxon signed ranks test of the CEC2013 large-scale benchmark suite}
	\label{table8}
	\begin{adjustbox}{center}
	\scalebox{0.6}{
		\begin{tabular}{|p{1.8cm}<{\centering}|p{1.5cm}<{\centering}|p{1.5cm}<{\centering}|p{1.5cm}<{\centering}|p{1.5cm}<{\centering}|p{1.5cm}<{\centering}|p{1.5cm}<{\centering}|p{2.2cm}<{\centering}|p{1.5cm}<{\centering}|p{1.5cm}<{\centering}|p{1.5cm}<{\centering}|}
			\hline
			\diagbox[width=6.45em]{Stat.}{SpadePSO \\ VS.}  & \tabincell{c}{PSO\\(1995)} & \tabincell{c}{CLPSO\\(2006)}  & \tabincell{c}{OLPSO\\(2011)} & \tabincell{c}{HCLPSO\\(2015)}   &\tabincell{c}{GL-PSO\\(2016)} & \tabincell{c}{TSLPSO\\(2019)} & \tabincell{c}{SPA-CatlePSO\\(2019)}& \tabincell{c}{CC-RDG3\\(2019)} & \tabincell{c}{XPSO\\(2020)} & \tabincell{c}{DMO\\(2022)}\\ \hline
			
			+  & 11 & 12 & 10 & 9 & 11 & 10 & 11 &6 &11 &15 \\ \hline
			-  & 4 & 3 & 5 & 6 & 4 & 5 & 4 &9 &4 &0\\ \hline
     $\approx$ & 0 & 0 & 0 & 0 &0  &0  &0 &0   &0 &0                             \\ \hline                         
			$p$-value & 0.04 & 0.04   & 0.07    & 0.23  & 0.03 & 0.03 &0.23 & 0.25 & 0.04 &0.00 \\ \hline
			
		\end{tabular}} 
	\end{adjustbox}
	\end{table}

\subsection{Experimental results on CEC2018 dynamic multi-objective optimization benchmark suite}
\label{sec4.5}

Dynamic optimization problems are challenging but critical, requiring an algorithm to respond to evolving environmental changes within a certain time period  (usually short). To assess the performance of SpadePSO on real-time, dynamic optimization problems, we conduct experiments on the CEC2018 dynamic multi-objective optimization benchmark suite, whose $D$ is 10. The benchmark suite consists of fourteen test functions: nine bi-objective (F1 $ \sim $ F9) and five tri-objective (F10 $ \sim $ F14) functions. For the CEC2018 dynamic multi-objective optimization benchmark suite, the environment changes after every $\tau_t$ iterations, i.e., the global optimal solution changes after every $\tau_t$ iterations. In our experiment, $\tau_t$ is set to 10, i.e., fast-changing environments. Therefore, we set the maximum number of iterations per run to 100$\cdot \tau_t$ and adopt the modified version of Inverted Generational Distance (MIGD) \citep{Zhou_A_2014} for performance evaluations. Specifically, we average the fitness of each algorithm in all environments. The formula of MIGD is as follows:
\begin{equation}
 \textit{MIGD} = \frac{1}{T} {\sum_{t=1}^{T} {\textit{IGD}(P_t^*,P_t)}} = \frac{1}{T} {\sum_{t=1}^{T} {\sum_{j=1}^{n_{P_t}} \frac{d_t^j}{n_{P_t}}}},
\end{equation}
where $P_t$ and $P_t^*$ denote a set of uniformly distributed points in the true Pareto Front (PF) and an approximation of the PF at time $t$, respectively, ${n_{P_t}} = |P_t|$, $d_t^j$ denotes the Euclidean distance between the $j$th member in $P_t$ and its nearest member in $P_t^*$, and $T$ is a set of discrete time points. We present the Mean and Std of fitness values of all algorithms for comparison in Table~\ref{table9}. Furthermore, because certain algorithms evaluate the fitness of many candidate solutions in each iteration, we also list the gaps of all algorithms. The gap is defined as follows:
\begin{equation}
\textit{gap} =  \frac{g}{g_{min}} -1,
\end{equation}
where $g$ denotes the number of fitness function evaluations of an algorithm, and $g_{min}$ denotes the minimum number of fitness function evaluations across all algorithms. In Table~\ref{table10}, we present the Wilcoxon signed ranks test of SpadePSO and all benchmarking algorithms.

\begin{table}[!t]
			\centering
			\caption{Comparison of SpadePSO with other algorithms on the CEC2018 dynamic multi-objective optimization benchmark suite}
			\label{table9}
			\scalebox{0.45}{
				\begin{tabular}{|c|c|c|c|c|c|c|c|c|c|c|c|}
					\hline
					\tabincell{c}{Function}          & \diagbox[width=6em]{Stat.}{Algorithm} & \tabincell{c}{PSO\\(1995)}     & \tabincell{c}{CLPSO\\(2006)}  & \tabincell{c}{OLPSO\\(2011)} & \tabincell{c}{HCLPSO\\(2015)}   &\tabincell{c}{GL-PSO\\(2016)} & \tabincell{c}{TSLPSO\\(2019)} & \tabincell{c}{SPA-CatlePSO\\(2019)}& \tabincell{c}{XPSO\\(2020)}  & \tabincell{c}{DMO\\(2022)}   & \tabincell{c}{SpadePSO\\(ours)} \\ \hline
\multirow{2}{*}{F1}  & Mean & 3.60E-01 & 3.82E-01 & 3.72E-01 & 3.72E-01 & 3.60E-01 & 3.56E-01 & 3.54E-01 & 3.60E-01 & 3.60E-01 & 3.53E-01 \\ \cline{2-12} 
                     & Std  & 5.87E-06 & 2.90E-03 & 6.46E-03 & 7.21E-03 & 3.99E-06 & 1.39E-03 & 2.86E-03 & 2.86E-07 & 1.33E-04 & 3.62E-03 \\ \hline
\multirow{2}{*}{F2}  & Mean & 3.19E-01 & 3.36E-01 & 3.27E-01 & 3.27E-01 & 3.19E-01 & 3.20E-01 & 3.23E-01 & 3.19E-01 & 3.20E-01 & 3.23E-01 \\  \cline{2-12} 
                     & Std  & 1.74E-05 & 3.41E-03 & 2.05E-03 & 1.99E-03 & 3.78E-07 & 4.23E-04 & 1.96E-04 & 1.53E-07 & 1.36E-04 & 5.49E-04 \\ \hline
\multirow{2}{*}{F3}  & Mean & 3.67E-01 & 4.30E-01 & 4.07E-01 & 3.96E-01 & 3.62E-01 & 3.63E-01 & 3.50E-01 & 3.63E-01 & 3.67E-01 &  3.48E-01\\
\cline{2-12} 
                     & Std  & 4.67E-04 & 2.32E-02 & 2.15E-02 & 1.18E-02 & 1.13E-04 & 9.27E-04 & 3.21E-03 & 5.35E-04 & 2.22E-03 & 5.78E-03 \\ \hline
\multirow{2}{*}{F4}  & Mean & 4.25E-01 & 6.00E-01 & 5.03E-01 & 4.96E-01 & 4.10E-01 & 4.21E-01 & 4.17E-01 & 3.85E-01 & 4.21E-01 & 4.14E-01\\
\cline{2-12} 
                     & Std  & 1.72E-03 & 4.12E-02 & 3.56E-02 & 1.12E-02 & 1.93E-02 & 6.19E-05 & 7.93E-03 & 2.56E-04 & 1.85E-02 & 9.07E-03 \\ \hline
\multirow{2}{*}{F5}  & Mean & 3.55E-01 & 3.85E-01 & 3.64E-01 & 3.63E-01 & 3.54E-01 & 3.54E-01 & 3.59E-01 & 3.54E-01 & 3.55E-01 & 3.59E-01 \\
\cline{2-12} 
                     & Std  & 1.55E-04 & 3.37E-03 & 2.60E-03 & 1.66E-03 & 1.03E-05 & 8.10E-05 & 1.15E-03 & 3.76E-05 & 2.96E-04 & 7.38E-04 \\ \hline
\multirow{2}{*}{F6}  & Mean & 1.53E+00 & 3.61E+00 & 2.98E+00 & 3.36E+00 & 2.91E-01 & 7.18E-01 & 1.42E+00 & 5.33E-01 & 3.19E-01 & 1.39E+00 \\
\cline{2-12} 
                     & Std  & 1.51E-01 & 1.68E-01 & 6.94E-01 & 4.25E-01 & 9.00E-02 & 2.64E-01 & 1.40E-01 & 2.20E-01 & 3.59E-02 & 1.11E-01 \\ \hline
\multirow{2}{*}{F7}  & Mean & 5.00E-01 & 5.15E-01 & 5.03E-01 & 5.04E-01 & 5.00E-01 & 5.00E-01 & 4.93E-01 & 4.89E-01 & 5.00E-01 & 4.93E-01 \\
\cline{2-12} 
                     & Std  & 1.07E-04 & 4.59E-03 & 1.32E-03 & 1.38E-03 & 4.29E-06 & 5.32E-05 & 1.55E-03 & 1.14E-05 & 1.78E-04 & 2.42E-03 \\ \hline
\multirow{2}{*}{F8}  & Mean & 3.45E-01 & 3.52E-01 & 3.46E-01 & 3.45E-01 & 3.45E-01 & 3.45E-01 & 3.43E-01 & 3.43E-01 & 3.52E-01 & 3.43E-01 \\
\cline{2-12} 
                     & Std  & 2.89E-05 & 3.42E-03 & 3.38E-04 & 1.71E-04 & 2.70E-02 & 7.97E-05 & 2.64E-04 & 4.44E-04 & 4.05E-03 & 1.15E-04 \\ \hline
\multirow{2}{*}{F9}  & Mean & 1.87E-01 & 3.91E-01 & 3.49E-01 & 2.54E-01 & 1.78E-01 & 1.82E-01 & 2.28E-01 & 1.83E-01 & 1.92E-01 & 2.14E-01  \\
\cline{2-12} 
                     & Std  & 4.39E-03 & 5.19E-02 & 3.93E-02 & 2.24E-02 & 5.04E-04 & 4.98E-03 & 8.94E-03 & 3.48E-03 & 1.04E-02 & 2.37E-02 \\ \hline
\multirow{2}{*}{F10} & Mean & 4.02E-01 & 4.48E-01 & 4.03E-01 & 4.31E-01 & 4.46E-01 & 4.10E-01 & 3.78E-01 & 4.26E-01 & 4.39E-01 & 3.77E-01 \\
\cline{2-12} 
                     & Std  & 4.15E-02 & 3.91E-03 & 4.36E-02 & 2.29E-02 & 1.41E-06 & 3.05E-02 & 1.32E-02 & 4.47E-02 & 7.02E-03 & 7.95E-03 \\ \hline
\multirow{2}{*}{F11} & Mean & 2.61E+01 & 2.60E+01 & 2.52E+01 & 2.55E+01 & 2.56E+01 & 2.50E+01 & 2.53E+01 & 2.54E+01 & 2.49E+01 & 2.53E+01 \\
\cline{2-12} 
                     & Std  & 4.14E-01 & 1.12E-01 & 3.65E-02 & 5.09E-02 & 1.44E-01 & 2.31E-02 & 3.89E-02 & 3.08E-01 & 2.25E-03 & 2.20E-02 \\ \hline
\multirow{2}{*}{F12} & Mean & 5.08E-01 & 5.65E-01 & 5.30E-01 & 5.25E-01 & 5.03E-01 & 5.05E-01 & 5.17E-01 & 5.04E-01 & 5.05E-01 & 5.18E-01 \\
\cline{2-12} 
                     & Std  & 1.34E-03 & 7.72E-03 & 4.24E-03 & 3.33E-03 & 1.68E-04 & 7.78E-04 & 2.42E-03 & 4.19E-04 & 5.97E-04 & 3.20E-03 \\ \hline
\multirow{2}{*}{F13} & Mean & 9.34E-01 & 9.47E-01 & 9.41E-01 & 9.38E-01 & 9.59E-01 & 9.36E-01 & 9.34E-01 & 9.32E-01 & 9.42E-01 & 9.34E-01 \\
\cline{2-12} 
                     & Std  & 1.24E-03 & 9.02E-03 & 2.57E-03 & 1.86E-03 & 1.11E-05 & 2.77E-03 & 4.24E-04 & 3.32E-04 & 4.41E-03 & 1.01E-03 \\ \hline
\multirow{2}{*}{F14} & Mean & 1.84E-01 & 2.04E-01 & 1.91E-01 & 1.89E-01 & 1.83E-01 & 1.83E-01 & 1.87E-01 & 1.83E-01 & 1.84E-01 & 1.86E-01 \\
\cline{2-12} 
                     & Std  & 1.84E-04 & 6.66E-03 & 1.99E-03 & 4.40E-04 & 5.66E-06 & 2.70E-05 & 3.01E-04 & 2.62E-05 & 1.75E-04 & 1.08E-03  \\ \hline
					& \pmb{gap}  & \pmb{0.00\%}   & \pmb{0.00\%}    & \pmb{263.11\%} & \pmb{0.0\%}   & \pmb{100.00\%} & \pmb{21.31\%}   &  \pmb{0.0\%} &  \pmb{0.0\%} & \pmb{70.49\%} &  \pmb{0.0\%}  \\ \hline
				\end{tabular}}
			\end{table}
			
As presented in Table~\ref{table9}, the performance of CLPSO, OLPSO, and HCLPSO is worse than that of PSO. There is a plausible reason that they only focus on exploration and ignore exploitation. In addition, for dynamic optimization problems, the response time for environmental changes is often tight. Therefore, the number of fitness function evaluations is vital for real-time applications. The gap of PSO, CLPSO, HCLPSO, SPA-CatlePSO, XPSO, and SpadePSO is 0.00\%, which means they are the best choices for real-time problems requiring immediate responses. As presented in Table~\ref{table10}, the difference between SpadePSO and PSO, GL-PSO, TSLPSO, XPSO, and DMO is not statistically significant. Comprehensively considering both the time and fitness value factors, XPSO is the algorithm that performs best on this benchmark suite, while SpadePSO is the second best. This also shows that there is still room to further extend our SpadePSO method beyond its current design to better to solve multi-objective dynamic problems.

\begin{table}[!t]
	\small
	\caption{Wilcoxon signed ranks test of the CEC2018 dynamic multi-objective optimization benchmark suite}
	\label{table10}
	\begin{adjustbox}{center}
	\scalebox{0.6}{
		\begin{tabular}{|p{1.8cm}<{\centering}|p{1.5cm}<{\centering}|p{1.5cm}<{\centering}|p{1.5cm}<{\centering}|p{1.5cm}<{\centering}|p{1.5cm}<{\centering}|p{1.5cm}<{\centering}|p{2.2cm}<{\centering}|p{1.5cm}<{\centering}|p{1.5cm}<{\centering}|}
			\hline
			\diagbox[width=6.45em]{Stat.}{SpadePSO \\ VS.}  & \tabincell{c}{PSO\\(1995)} & \tabincell{c}{CLPSO\\(2006)}  & \tabincell{c}{OLPSO\\(2011)} & \tabincell{c}{HCLPSO\\(2015)}   &\tabincell{c}{GL-PSO\\(2016)} & \tabincell{c}{TSLPSO\\(2019)} & \tabincell{c}{SPA-CatlePSO\\(2019)}& \tabincell{c}{XPSO\\(2020)} & \tabincell{c}{DMO\\(2022)}\\ \hline
			
			+  & 8 & 14 & 13 & 14 & 7 & 7 & 7 &4 &7 \\ \hline
			-  & 5 & 0  & 1  & 0  & 7 & 7 & 1  &9 &7\\ \hline
     $\approx$ & 0 & 0  & 0  & 0  &0  &0  &6   &1 &0                             \\ \hline                         
			$p$-value & 0.24 & 0.00   & 0.01    & 0.00  & 0.84 & 0.47 &0.04 & 0.54 &0.96 \\ \hline

		\end{tabular}} 
	\end{adjustbox}
	\end{table}

\subsection{Spread spectrum radar polyphase (SSRP) code design }
\label{sec4.6}

In this subsection, we conduct experiments to solve the real-world optimization problem of SSRP code design, whose $D$ is 20. As the problem is NP-hard and its fitness function is piecewise-smooth, SSRP code design has been widely used as the optimization objective of swarm intelligence algorithms \citep{mladenovic_solving_2003,das_differential_2009}.  Therefore, we select SSRP code design as a real-world problem  to compare the performance of various algorithms. We adopt the min-max nonlinear optimization problem model \citep{dukic_method_1990} as the fitness function, which is defined as follows:
\begin{equation}
\begin{aligned}
& \min_{x\in X}{f}(x)=\max \{{\phi_1}(x), ..., {\phi_{2m}}(x)\},\\
& X=\{(x_1,...,x_n) \in R^n |  0 \leq x_j \leq 2\pi, j=1,...,n\},\\
\end{aligned}
\end{equation}
where $m = 2n-1$, and

\begin{small}
\begin{equation}
\begin{aligned}
& {\phi_{2i-1}}(x)=\sum_{j=i}^n \cos \bigg{(}{\sum_{k=|{2i-j-1}|+1}^j x_k} \bigg{)}, i=1,...n,\\
& {\phi_{2i}}(x)=0.5+\sum_{j=i+1}^n \cos \bigg{(}{\sum_{k=|{2i-j}|+1}^j x_k} \bigg{)}, i=1,...n-1,\\
& {\phi_{m+i}}(x)=-{\phi_i}(x), i=1,...m.
\end{aligned}
\end{equation}
\end{small}

To better quantitatively assess the experimental results, we select the Mean and Std of fitness values and the one-tailed $t$-test results as the performance evaluation metrics. One-tailed $t$-test is performed on the mean value with freedom at the 0.1 level. As presented in Table~\ref{table11}, SpadePSO performs significantly better than CLPSO, OLPSO, HCLPSO, GLPSO, TSLPSO, and SPA-CatlePSO. Although the mean values of SpadePSO, PSO, and XPSO are not significantly different, the Std of SpadePSO is approximately half of PSO and XPSO, which demonstrates that SpadePSO is relatively more stable. 

%best  & 1.31 & 1.40 & 1.32 & 1.31 & 1.15 & 1.23 &1.03 &1.23\\ \hline
\begin{table}[!t]
	\small
	\caption{Comparison of SpadePSO with state-of-the-art PSO variants on the SSRP code design problem}
	\label{table11}
	\begin{adjustbox}{center}
	\scalebox{0.6}{
		\begin{tabular}{|p{1.8cm}<{\centering}|p{1.5cm}<{\centering}|p{1.5cm}<{\centering}|p{1.5cm}<{\centering}|p{1.5cm}<{\centering}|p{1.5cm}<{\centering}|p{1.5cm}<{\centering}|p{2.2cm}<{\centering}|p{1.5cm}<{\centering}|p{1.5cm}<{\centering}|p{1.5cm}<{\centering}|}
			\hline
			\diagbox[width=6.45em]{Stat.}{Algorithm}  & \tabincell{c}{PSO\\(1995)} & \tabincell{c}{CLPSO\\(2006)}  & \tabincell{c}{OLPSO\\(2011)} & \tabincell{c}{HCLPSO\\(2015)}   &\tabincell{c}{GL-PSO\\(2016)} & \tabincell{c}{TSLPSO\\(2019)} & \tabincell{c}{SPA-CatlePSO\\(2019)}& \tabincell{c}{XPSO\\(2020)} & \tabincell{c}{DMO\\(2022)} & \tabincell{c}{SpadePSO\\(ours)}\\ \hline
			
			Mean  & 1.53 & 1.73 & 1.91 & 1.58 & 1.83 & 1.65 & 1.74 &1.53 &1.61 &1.52 \\ \hline
			Std   & 0.18 & 0.10 & 0.25 & 0.15 & 0.18 & 0.14 & 0.12 &0.19 &0.06 &0.12\\ \hline
			$t$-test & $\approx$ & +   & +    & +  & + & + & + & $\approx$ & + & \\ \hline
			
		\end{tabular}} 
	\end{adjustbox}
	\end{table}
\subsection{Ordinary differential equations models inference}
\label{sec4.7}
Modeling dynamic systems in the field of physics, biology, and chemistry is commonly achieved by inferring the Ordinary Differential Equation (ODE) based on the observed time-series data \citep{xue_sieve_2010, heinonen_learning_2018}. Owing to the requirement of higher computational resources and the design of the solution space, inferring the structure and parameters of ODE models simultaneously is a challenging task \citep{usman_inferring_2020,heinonen_learning_2018}. Moreover, such problems normally have both discrete and continuous variables, making them more challenging. To demonstrate the effectiveness of SpadePSO, we simultaneously infer the structure and parameters of the HIV model \citep{perelson_mathematical_1999}, a well-known ODE instance, from scratch. Specifically, we do not know about the values of parameters and interrelation between variables but only know the number of variables and the raw time-series data. In our experiment, the time-series data are obtained by computing the HIV model \citep{perelson_mathematical_1999}, which is described as follows: 

\begin{equation}
\begin{aligned}
& \mathrm{d}C/\mathrm{d}t = 80 - 15 \cdot C - 0.00002 \cdot C \cdot V,\\
& \mathrm{d}I/\mathrm{d}t = 0.00002 \cdot C \cdot V - 0.55 \cdot I,\\
& \mathrm{d}V/\mathrm{d}t = 495 \cdot I - 0.5 \cdot V - 0.00002 \cdot C \cdot V,
\label{e27}
\end{aligned}
\end{equation}
where $C$, $I$, and $V$ denote the number of uninfected cells, the number of infected CD4+T lymphocytes, and the number of free viruses, respectively. The initial conditions (variable values at time $t = 0$) for computing the time-series data are as follows: $C_0 = 100$, $I_0 = 150$, and $V_0 = 50,000$ \citep{tian_latinpso_2019}. 

There are three equations in the HIV model. Tian et al. \cite{tian_latinpso_2019} defined the general form of these three equations as follows:
%Therefore, the HIV model can be represented using three general forms.

\begin{equation}
\mathrm{d}x_i/\mathrm{d}t = \pm k_1 \cdot x_i \pm k_2 \cdot x_a \pm k_3 \cdot x_b \cdot x_c \pm k_4,
\label{e28}
\end{equation}
where $k_1, k_2, k_3$, and $k_4$ denote the parameters of the equation, symbol $\pm$ denotes one of the addition or subtraction operators, and $x_i$, $x_a$, $x_b$, and $x_c$ denote the variables used in the model. In addition, $i, b, c \in \{1,2,3\}$ and $a \in \{1,2,3\} \setminus \{i\}$. For example, if $i=1$, $x_a$ represents one of the variables $x_2$ and $x_3$, where $x_b$ and $x_c$ represent one of the variables $x_1$, $x_2$ and $x_3$. %In addition, this general form assumes that the equation of the HIV model is combined by four different items, including two single-variable items, one two-variable item and one item without variable. More items or multi-variable items such as three and so on,

The first variable $x_i$ must exist in Eq. \eqref{e28} because Eq. \eqref{e28} is a differential equation about $x_i$, so we do not code $x_i$ into the corresponding PSO representation. Therefore, the representation of Eq. \eqref{e28} is formulated as $ \{ k_1, k_2, k_3, k_4,(\pm, \pm, \pm, \pm, x_a, x_b, x_c) \}$, and there are 192 possible structures for each general form as presented in Table~\ref{table12}. To solve the ODE inference problem,  each possible structure is represented by an index as a dimension of the problem. Therefore, the structure of an equation can be regarded as a dimension of the problem, and parameters $k_1, k_2, k_3$, and $ k_4$ are regarded as four dimensions of the problem. As the HIV model has three equations, the dimension of the problem is fifteen, out of which twelve dimensions represent the parameters and three dimensions represent the structure of three equations. 

\begin{table}[!t]
	\centering
	\caption{Enumeration of ODE structures of Eq. \eqref{e28}, when $i=1$}
	\label{table12}
	\begin{tabular}{llll}
		\toprule
		Serial number & ODE structures  \\
		\midrule
		$1$ & $ (+, +, +, +, x_2, x_1, x_1)$\\
		$2$ & $ (+, +, +, +, x_2, x_1, x_2)$\\
		$3$ & $ (+, +, +, +, x_2, x_1, x_3)$\\
		$4$ & $ (+, +, +, +, x_2, x_2, x_2)$\\
		$5$ & $ (+, +, +, +, x_2, x_2, x_3)$\\
		$6$ & $ (+, +, +, +, x_2, x_3, x_3)$\\
		$7$ & $ (+, +, +, +, x_3, x_1, x_1)$\\
		$...$ & $......$\\
		$13$ & $ (+, +, +, -, x_2, x_1, x_1)$\\
		$...$ & $......$\\
		$192$ & $ (-, -, -, -, x_2, x_3, x_3)$\\
		\bottomrule
	\end{tabular}

\end{table}

The fitness function of particles is defined as the squared error between the time-series data computed by the inferred HIV model using PSO variants and the given target time-series data, which can be expressed as follows:

\begin{equation}
f(\textbf{x}) = \sum_{i=1}^U{\sum_{s=0}^H{(\textbf{x}_i^\prime{(h_0+s\Delta h))}-(\textbf{x}_i{(h_0+s\Delta h))}}^2}, 
\label{e29}
\end{equation}
where $s \in \{0,1,2,…,H-1\}$, $h_0$ denotes the starting time, $\Delta h$ denotes the step size, $U$ denotes the number of the state variables, and $H$ denotes the number of data points. The term $\textbf{x}_i(h_0+s\Delta h)$ denotes the given target time-series data, and the term $\textbf{x}_i^\prime (h_0+s\Delta h)$ denotes the time-series data computed by the inferred HIV model using PSO variants. If the inferred model is identical to the HIV model, the fitness value is 0. The larger the fitness value, the higher the difference between the inferred model and the HIV model. 

Table~\ref{table13} presents the Mean and Std of fitness values of PSO variants, as well as the one-tailed $t$-test results. The one-tailed $t$-test results show that SpadePSO performs significantly better than PSO, HCLPSO, GL-PSO, TSLPSO, and SPA-CatlePSO. The mean values of CLPSO, OLPSO, XPSO, and SpadePSO are not significantly different, but the Std of SpadePSO is approximately half of CLPSO, OLPSO, and XPSO. Thus, SpadePSO is shown to be relatively more stable.

\begin{table}[!t]
\small
				\centering
				\caption{Comparison of SpadePSO with the state-of-the-art PSO variants on ODE models inference}
				\label{table13}
				\begin{adjustbox}{center}
				\scalebox{0.6}{
		\begin{tabular}{|p{1.8cm}<{\centering}|p{1.5cm}<{\centering}|p{1.5cm}<{\centering}|p{1.5cm}<{\centering}|p{1.5cm}<{\centering}|p{1.5cm}<{\centering}|p{1.5cm}<{\centering}|p{2.2cm}<{\centering}|p{1.5cm}<{\centering}|p{1.5cm}<{\centering}|p{1.5cm}<{\centering}|}
			\hline
			\diagbox[width=6.45em]{Stat.}{Algorithm}  & \tabincell{c}{PSO\\(1995)} & \tabincell{c}{CLPSO\\(2006)}  & \tabincell{c}{OLPSO\\(2011)} & \tabincell{c}{HCLPSO\\(2015)}   &\tabincell{c}{GL-PSO\\(2016)} & \tabincell{c}{TSLPSO\\(2019)} & \tabincell{c}{SPA-CatlePSO\\(2019)} & \tabincell{c}{XPSO\\(2020)} & \tabincell{c}{DMO\\(2022)} & \tabincell{c}{SpadePSO\\(ours)}\\ \hline
			
			Mean  & 8.91E+03 &6.49E+03  & 5.92E+03  & 7.76E+03  & 7.59E+03  & 7.43E+03 & 7.04E+03  &6.21E+03 &7.40E+03 &6.58E+03  \\ \hline
			Std   & 2.31E+03 & 3.32E+03  & 3.95E+03  & 2.72E+03  & 1.25E+03  & 1.96E+03 &2.78E+03 &3.48E+03 &4.83E+02 &1.62E+03 \\ \hline
			$t$-test & +& $\approx$   & $\approx$    & +   & + & + & + & $\approx$ & + & \\ \hline
			
		\end{tabular}} 
		\end{adjustbox}
	\end{table}

\section{Conclusion}
\label{sec5}
We propose an adaptive topology based on Euclidean distance and incorporate it along with the surprisingly popular algorithm into the novel  SpadePSO model. We conduct experiments on three benchmark suites and two real-world optimization problems. The experimental results show that SpadePSO outperforms the state-of-the-art PSO variants on the full CEC2014 benchmark suite, CEC2013 large-scale benchmark suite, and two real-world applications. 

Through the introduction of Linear Population Size Reduction, L-SHADE, a DE variant, performs significantly better than PSO variants with a relatively large population size. We plan to introduce Linear Population Size Reduction into SpadePSO for further performance improvement.

%In short, experiments show the effectiveness of SPA, and the metric based on SPA to evaluate particles can also be applied to other algorithms.

\section{Acknowledgement}

This work is supported by the National Natural Science Foundation of China (61972174 and 61972175), the Jilin Natural Science Foundation (20200201163JC), the Guangdong Science and Technology Planning Project (2020A0505100018), Guangdong Universities' Innovation Team Project (2021KCXTD015), and Guangdong Key Disciplines Project (2021ZDJS138).

  \appendix
  \renewcommand{\appendixname}{Appendix}

  \section{}
In Table \ref{table14}, we show the Best, Mean and Std of fitness values obtained by SpadePSO on the CEC2014 benchmark suite.
			
			\begin{table}[!t]
				
				\caption{Results of SpadePSO on the CEC2014 benchmark suite (D = 10, 30, 50 and 100)}
				\label{table14}
				\begin{adjustbox}{center}
				\scalebox{0.7}{
					\begin{tabular}{|c|c|c|c|c|c|c|c|c|c|c|c|c|}
						\hline
						\multicolumn{1}{|c|}{{Dim.}} & \multicolumn{3}{c|}{10D}                                                                            & \multicolumn{3}{c|}{30D}   & \multicolumn{3}{c|}{50D} & \multicolumn{3}{c|}{100D} \\ \hline
						\multicolumn{1}{|c|}{\diagbox[width=4em]{Func.}{Stat.}}                       & \multicolumn{1}{c|}{Best}       & \multicolumn{1}{c|}{Mean}       & \multicolumn{1}{c|}{Std}        & \multicolumn{1}{c|}{Best}       & \multicolumn{1}{c|}{Mean}       & \multicolumn{1}{c|}{Std}   & \multicolumn{1}{c|}{Best}       & \multicolumn{1}{c|}{Mean}       & \multicolumn{1}{c|}{Std} & \multicolumn{1}{c|}{Best}       & \multicolumn{1}{c|}{Mean}       & \multicolumn{1}{c|}{Std}     \\ \hline
						{1}                          & {7.75E+01} & {1.09E+04} & {1.34E+04} & {2.79E+04} & {2.11E+05} & {1.52E+05} & {3.43E+05} & {6.31E+05} & {2.45E+05} & {1.39E+06} & {3.05E+06} & {8.78E+05}\\
						{2}                          & {5.32E-02} & {3.74E+01} & {5.00E+01} & {6.89E-05} & {1.11E+01} & {2.69E+02} & {3.58E+00} & {1.70E+02} & {3.33E+02} & {3.29E+00} & {5.62E+02} & {8.03E+02}\\
						{3}                          & {9.59E-03} & {6.43E+01} & {9.44E+01} & {3.61E-01} & {1.28E+02} & {1.28E+02} &{1.37E+02} & {1.68E+03} & {9.72E+02} & {2.68E+02} & {1.89E+03} & {1.30E+03}\\
						{4}                          & {2.05E-03} & {1.44E+00} & {3.71E+00} & {3.30E-02} & {3.91E+01} & {3.26E+01} & {2.17E+01} & {8.59E+01} & {2.01E+01} & {1.11E+02} & {1.98E+02} & {3.67E+01}\\
						{5}                          & {0.00E+00} & {1.85E+01} & {5.40E+00} & {2.01E+01} & {2.02E+01} & {3.92E-02} & {2.00E+01} & {2.03E+01} & {9.49E-02} & {2.00E+01} & {2.02E+01} & {1.68E-01}\\
						{6}                          & {1.13E-04} & {1.25E-02} & {2.17E-02} & {6.26E-01} & {2.11E+00} & {1.10E+00} & {2.31E+00} & {9.46E+00} & {3.04E+00} & {4.03E+01} & {5.55E+01} & {6.13E+00}\\
						{7}                          & {3.25E-03} & {4.22E-02} & {2.28E-02} & {0.00E+00} & {2.87E-04} & {1.11E-03} & {0.00E+00} & {5.76E-03} & {7.24E-03} & {0.00E+00} & {1.39E-03} & {3.85E-03}\\
						{8}                          & {0.00E+00} & {0.00E+00} & {0.00E+00} & {0.00E+00} & {0.00E+00} & {0.00E+00} & {0.00E+00} & {0.00E+00} & {0.00E+00} & {0.00E+00} & {5.58E-02} & {2.34E-01}\\
						{9}                          & {1.53E+00} & {4.48E+00} & {1.68E+00} & {1.59E+01} & {4.35E+01} & {1.12E+01} & {3.98E+01} & {9.22E+01} & {2.33E+01} & {1.84E+02} & {2.79E+02} & {3.66E+01}\\
						{10}                         & {0.00E+00} & {1.84E-01} & {6.55E-01} & {4.17E-02} & {4.91E-01} & {8.01E-01} & {1.13E-01} & {1.22E+01} & {3.52E+01} & {2.52E-01} & {1.74E+01} & {4.69E+01}\\
						{11}                         & {1.87E+01} & {1.60E+02} & {1.21E+02} & {1.17E+03} & {1.91E+03} & {3.15E+02} & {3.07E+03} & {4.22E+03} & {4.34E+02} & {9.30E+03} & {1.11E+04} & {7.00E+02}\\
						{12}                         & {5.60E-02} & {2.10E-01} & {5.90E-02} & {5.51E-02} & {2.39E-01} & {5.83E-02} & {1.03E-01} & {2.13E-01} & {5.82E-02} & {1.75E-01} & {2.87E-01} & {7.33E-02}\\
						{13}                         & {2.54E-02} & {8.67E-02} & {3.37E-02} & {1.09E-01} & {2.27E-01} & {6.27E-02} & {2.22E-01} & {3.30E-01} & {5.84E-02} & {2.84E-01} & {4.13E-01} & {5.10E-02} \\
						{14}                         & {2.61E-02} & {8.26E-02} & {3.43E-02} & {1.66E-01} & {2.36E-01} & {3.11E-02} & {1.94E-01} & {2.69E-01} & {2.91E-02} & {2.50E-01} & {3.15E-01} & {2.60E-02}\\
						{15}                         & {3.99E-01} & {7.66E-01} & {2.15E-01} & {1.52E+00} & {4.32E+00} & {1.53E+00} & {4.84E+00} & {9.05E+00} & {2.20E+00} & {2.19E+01} & {3.69E+01} & {8.04E+00}\\
						{16}                         & {2.07E-01} & {1.32E+00} & {4.28E-01} & {7.30E+00} & {9.36E+00} & {6.76E-01} & {1.42E+01} & {1.77E+01} & {9.49E-01} & {3.89E+01} & {4.05E+01} & {6.24E-01}\\
						{17}                         & {1.88E+01} & {8.49E+02} & {7.01E+02} & {1.49E+03} & {8.82E+04} & {5.61E+04} & {1.83E+04} & {1.32E+05} & {1.07E+05} & {2.64E+05} & {8.04E+05} & {3.24E+05}\\
						{18}                         & {4.24E+00} & {3.02E+02} & {6.94E+02} & {3.19E+01} & {1.49E+02} & {1.36E+02} & {6.28E+01} & {1.52E+02} & {5.10E+01} & {2.51E+02} & {3.99E+02} & {1.48E+02}\\
						{19}                         & {3.91E-02} & {5.64E-01} & {3.66E-01} & {2.84E+00} & {4.71E+00} & {1.12E+00} & {8.37E+01} & {1.34E+01} & {3.01E+00} & {2.94E+01} & {7.41E+01} & {1.81E+01}\\
						{20}                         & {2.44E+00} & {3.07E+01} & {4.62E+01} & {1.05E+02} & {8.56E+02} & {5.91E+02} & {3.44E+02} & {1.29E+03} & {8.34E+02} & {1.62E+03} & {4.50E+03} & {1.94E+03}\\
						{21}                         & {6.62E+00} & {8.53E+01} & {8.13E+01} & {3.51E+03} & {3.86E+04} & {3.35E+04} & {2.43E+04} & {1.72E+05} & {1.68E+05} & {1.29E+05} & {5.03E+05} & {3.74E+05}\\
						{22}                         & {1.98E-02} & {1.94E+00} & {5.12E+00} & {2.81E+01} & {1.91E+02} & {7.33E+01} & {1.59E+02} & {6.93E+02} & {1.85E+02} & {1.12E+03} & {1.88E+03} & {3.13E+02}\\
						{23}                         & {3.29E+02} & {3.29E+02} & {2.59E-13} & {3.15E+02} & {3.15E+02} & {1.90E-12} & {3.44E+02} & {3.44E+02} & {1.03E-12} & {3.48E+02} & {3.48E+02} & {2.93E-11}\\
						{24}                         & {1.00E+02} & {1.12E+02} & {2.93E+00} & {2.24E+02} & {2.25E+02} & {1.12E+00} & {2.56E+02} & {2.60E+02} & {3.60E+00} & {3.57E+02} & {3.62E+02} & {1.98E+00}\\
						{25}                         & {1.04E+02} & {1.31E+02} & {2.18E+01} & {2.03E+02} & {2.06E+02} & {1.50E+00} & {2.07E+02} & {2.14E+02} & {2.76E+00} & {2.46E+02} & {2.56E+02} & {5.10E+00}\\
						{26}                         & {1.00E+02} & {1.00E+02} & {2.96E-02} & {1.00E+02} & {1.00E+02} & {5.51E-02} & {1.00E+02} & {1.06E+02} & {2.24E-02} & {1.00E+02} & {1.96E+02} & {1.94E-01}\\
						{27}                         & {9.45E-01} & {4.44E+01} & {1.06E+02} & {3.45E+02} & {3.97E+02} & {1.52E+01} & {4.25E+02} & {5.45E+02} & {8.30E+02} & {5.09E+02} & {1.54E+03} & {4.88E+02}\\
						{28}                         & {3.44E+02} & {3.75E+02} & {2.25E+01} & {7.66E+02} & {8.88E+02} & {4.81E+01} & {1.17E+03} & {1.52E+03} & {2.01E+02} & {3.46E+03} & {5.26E+03} & {7.65E+02}\\
						{29}                         & {2.33E+02} & {2.79E+02} & {3.56E+01} & {7.95E+02} & {9.52E+02} & {1.05E+02} & {9.27E+02} & {1.21E+03} & {1.85E+02} & {1.36E+03} & {1.74E+03} & {3.06E+02}\\
						{30}                         & {4.96E+02} & {6.58E+02} & {1.01E+02} & {1.11E+03} & {1.87E+03} & {4.42E+02}  & {8.55E+03} & {1.05E+04} & {9.50E+02} & {7.87E+03} & {9.68E+03} & {8.32E+02}\\ \hline
					\end{tabular}}
					\end{adjustbox}
				\end{table}
\small
%\end{appendices}
%\section*{References}
\clearpage
\bibliography{SpadePSO}
%\end{sloppypar}
\end{document}